\documentclass{article}

\PassOptionsToPackage{numbers, compress}{natbib}

\usepackage[preprint]{neurips_2024}

\usepackage[utf8]{inputenc} 
\usepackage[T1]{fontenc}    
\usepackage[dvipsnames,table,xcdraw]{xcolor}

\usepackage[pagebackref=false, breaklinks=true, letterpaper=true, colorlinks,
            citecolor=citecolor, linkcolor=linkcolor, bookmarks=false]{hyperref}
\definecolor{citecolor}{HTML}{0071BC}
\definecolor{linkcolor}{HTML}{ED1C24}

\usepackage{url}            
\usepackage{booktabs}       
\usepackage{amsfonts}       
\usepackage{nicefrac}       
\usepackage{microtype}      
\usepackage{xcolor}         
\usepackage{algpseudocode}
\usepackage{bm}
\usepackage{mathtools}
\usepackage{algorithm}
\usepackage{algpseudocode}
\usepackage{lipsum} 
\usepackage{placeins} 

\title{MotionFollower: Editing Video Motion via Lightweight Score-Guided Diffusion}

\author{Shuyuan Tu$^{1,2}$ \ \ 
Qi Dai$^{3}$ \ \
Zihao Zhang$^{1,2}$ \ \
Sicheng Xie$^{1,2}$ \ \
Zhi-Qi Cheng$^{4}$ \AND
Chong Luo$^3$ \ \ 
Xintong Han$^5$ \ \ 
Zuxuan Wu$^{1,2}$ \ \ 
Yu-Gang Jiang$^{1,2}$ \vspace{0.05in}\\
{$^1$Shanghai Key Lab of Intell. Info. Processing, School of CS, Fudan University} \\
{$^2$Shanghai Collaborative Innovation Center of Intelligent Visual Computing} \\
{$^3$Microsoft Research Asia}  \quad \quad 
{$^4$Carnegie Mellon University} \quad  \quad 
{$^5$Huya Inc.} \\
{\url{https://francis-rings.github.io/MotionFollower/}}
}

\begin{document}

\maketitle

\begin{figure}[h]
\begin{center}
\includegraphics[width=0.98\linewidth, trim=2.3cm 0.1cm 0cm 0cm, clip]{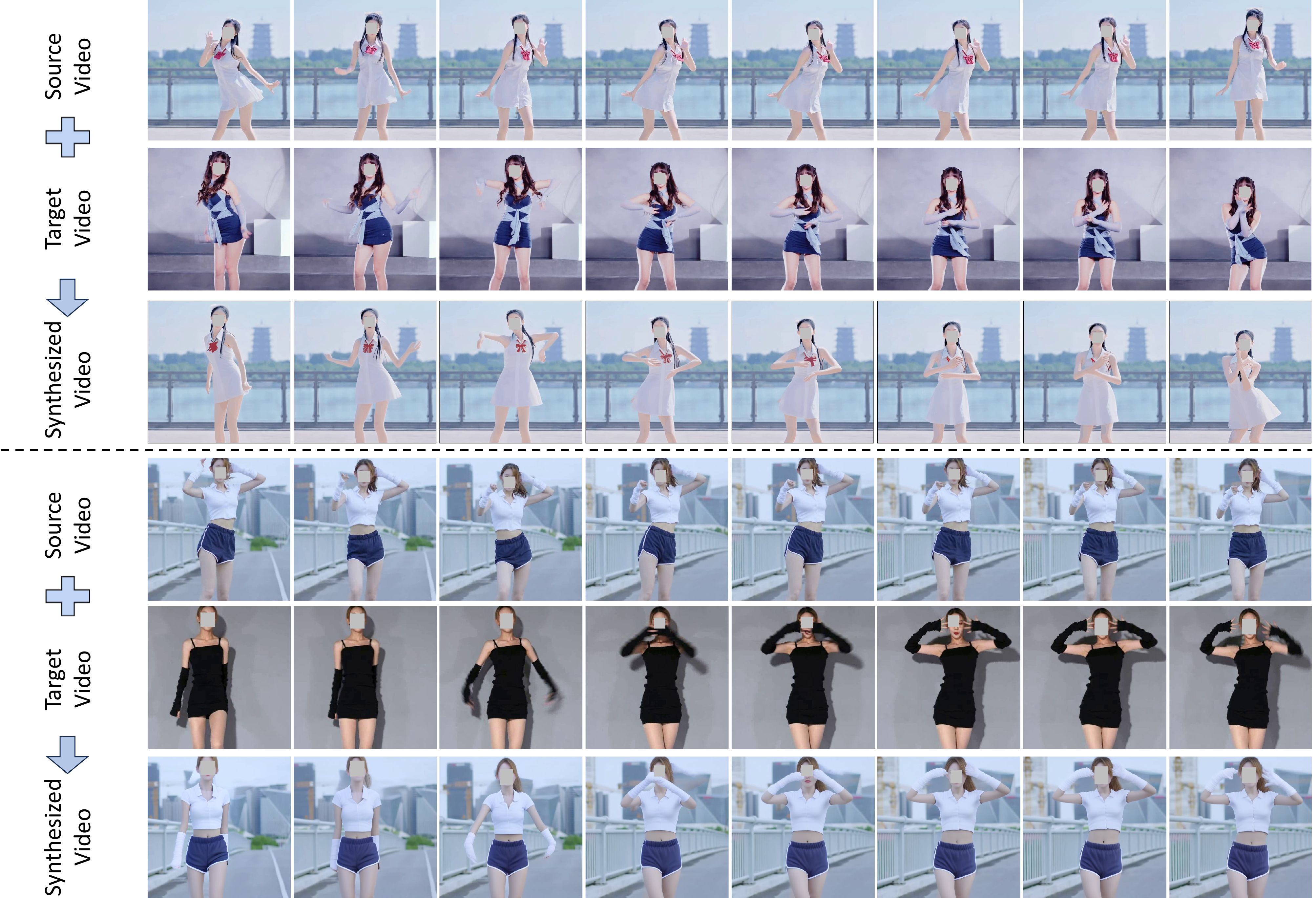}
\end{center}
\vspace{-0.4cm}
   \caption{\small MotionFollower: a lightweight motion editing method for transferring motion from target video to source while keeping source background, protagonists' appearance, and camera movement.}
\label{fig:general_case}
\vspace{-0.1cm}
\end{figure}

\begin{abstract}
Despite impressive advancements in diffusion-based video editing models in altering video attributes, there has been limited exploration into modifying motion information while preserving the original protagonist's appearance and background. In this paper, we propose MotionFollower, a lightweight score-guided diffusion model for video motion editing. To introduce conditional controls to the denoising process, MotionFollower leverages two of our proposed lightweight signal controllers, one for poses and the other for appearances, both of which consist of convolution blocks without involving heavy attention calculations. Further, we design a score guidance principle based on a two-branch architecture, including the reconstruction and editing branches, which significantly enhance the modeling capability of texture details and complicated backgrounds. 
Concretely, we enforce several consistency regularizers and losses during the score estimation.
The resulting gradients thus inject appropriate guidance to the intermediate latents, forcing the model to preserve the original background details and protagonists' appearances without interfering with the motion modification.
Experiments demonstrate the competitive motion editing ability of MotionFollower qualitatively and quantitatively. Compared with MotionEditor, the most advanced motion editing model, MotionFollower achieves an approximately 80\% reduction in GPU memory while delivering superior motion editing performance and exclusively supporting large camera movements and actions.
\vspace{-0.15cm}
\end{abstract}

\section{Introduction}
Diffusion models~\cite{dhariwal2021diffusion,ho2020denoising,ho2022cascaded,nichol2021improved,song2020denoising,rombach2022high,meng2021sdedit,mou2023dragondiffusion} have demonstrated their superiority in performing image and video generation, which promote numerous studies in video editing~\cite{wu2023tune,qi2023fatezero,ceylan2023pix2video,khachatryan2023text2video,wang2023zero,cong2023flatten,lee2023soundini,feng2023ccedit}. While considerable progress has been made, current models predominantly focus on attribute-level editing, such as video style transferring and manipulation for background and protagonist's appearance. Despite being recognized as one of the most distinct and sophisticated features compared to images, motion information is largely overlooked in existing models. Inspired by MotionEditor~\cite{tu2024motioneditor}, we aim to manipulate the motion of a video in alignment with a target video, which is regarded as a higher-level and more challenging video editing scenario---motion editing. In particular, given a target video and a source video, we wish to transfer the motion information of the target video to the protagonist in the source while preserving the original details.

As of current literature, MotionEditor~\cite{tu2024motioneditor} is the pioneer work capable of editing motions, leveraging an adapter appended to ControlNet~\cite{zhang2023adding} and an attention injection mechanism. However, it presents a degrading performance in terms of temporal consistency and suffers from substantial computation costs stemming from the ControlNet and the attention injection. Further, the performance of MotionEditor tends to deteriorate when encountering specific scenarios, such as large camera movements and complex backgrounds. 
Recent studies have also explored human motion transfer~\cite{liu2019liquid,siarohin2021motion,hu2023animate,wang2023disco,xu2023magicanimate,zhu2024champ} and pose-conditioned video generation~\cite{ma2023follow,zhang2023controlvideo}. 
The former aims to animate a given image based on target signals, while the latter tends to generate signal-guided videos without preserving a desired appearance. 
Video motion editing has a considerable discrepancy, primarily manipulating the video's motion while maintaining other extraneous details, such as camera movements, per-frame background variations, and the protagonist's appearance.

\begin{figure*}[t!]
\begin{center}
\includegraphics[width=\linewidth]{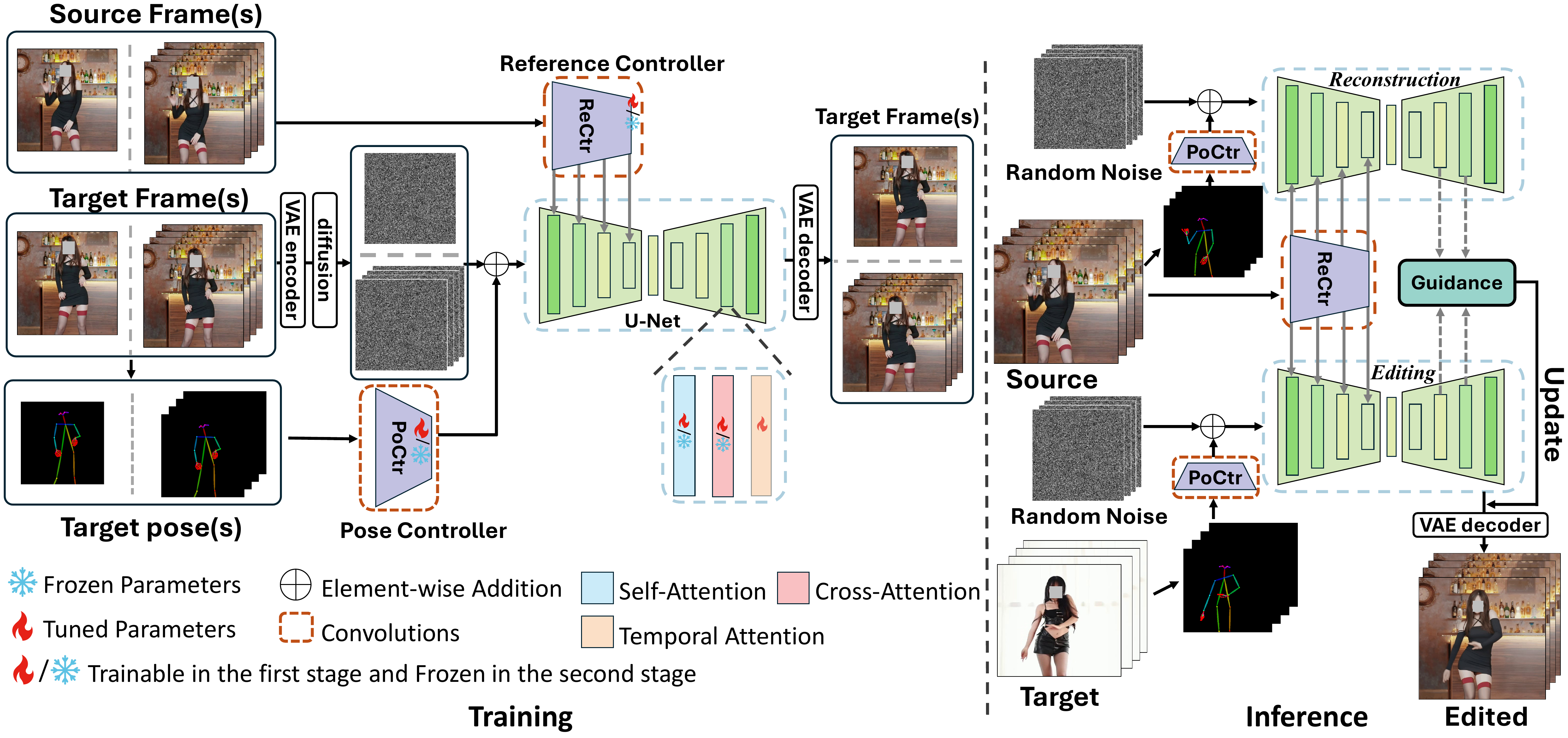}
\end{center}
\vspace{-0.3cm}
   \caption{\small Architecture overview. In training, two lightweight signal controllers and U-Net are trainable. 
   The model is first trained with single frame (first stage), then with video clip (second stage).
   In inference, we build a two-branch structure, one for reconstruction and the other for editing.
   Score guidance is computed using features from the two branches, which is then used to update the latent.
   }
\label{fig:overview}
\vspace{-0.4cm}
\end{figure*}

We propose a novel editing approach, MotionFollower, to tackle the above issues as summarized in Fig. \ref{fig:overview}. We design two lightweight signal controllers (Pose Controller and Reference Controller), to enhance capabilities for controlling pose and modeling appearance. These two controllers comprise pure convolutions without costly attention calculations, considerably reducing the computations. 

We further propose a novel score-guided principle that impels the model to keep the original semantic details during inference, \emph{e.g.} background and camera movement.
Instead of injecting attention maps or interacting with source key/value as in previous methods, which may cause unexpected overlapping and shadow flickering, we introduce score regularization to improve the consistency between source and edited video.
The denoising process can be regarded as a continuous process~\cite{song2020score}, which can be described as the score function. This score function steers the denoising towards a specific direction. Thus, we can influence the model to denoise data according to our preferences by incorporating additional regularization and hence ensure the outputs retain the essential semantic details of the source video \emph{e.g.}, background and camera movement.
Specifically, with a reconstruction branch and an editing branch, we utilize segmentation masks to calculate multiple losses corresponding to regions of different granularity levels.
Guided by the score function, the latent from denoising U-Net is then optimized by the loss gradients, facilitating regional consistency between the editing and reconstruction branches. Our guidance principle does not update the weights of the trained model.

In conclusion, our contributions are as follows: (1) We introduce two lightweight signal controllers to replace the heavy ControlNet, without losing control ability and performance.
(2) We propose a novel score guidance principle, enabling the diffusion model to preserve regional details of the source by introducing several regularizations into the score function, thereby increasing content consistency.
(3) MotionFollower achieves an 80\% reduction in GPU memory compared to the leading competitor, MotionEditor, while demonstrating superior performance, particularly with specific videos featuring large camera movements and complex backgrounds.


\section{Method}
As illustrated in Fig. \ref{fig:overview}, we build MotionFollower based on a T2I diffusion model (\emph{i.e.} LDM~\cite{rombach2022high}).
We improve the U-Net architecture by integrating motion modules (temporal layers) \cite{guo2023animatediff}, inflating it to a 3D U-Net.
To support conditional signals, we propose two lightweight conditional controllers: a Pose Controller (PoCtr) and a Reference Controller (ReCtr), which encode the target poses and source appearance, respectively.
Compared to ControlNet~\cite{zhang2023adding}, our controllers are more parameter-efficient, using only a few convolution layers instead of self- or cross-attention layers.

During inference, MotionFollower transfers the motion from the given target video to the source video while preserving the protagonist's appearance and the background.
This is achieved through a two-branch architecture, consisting of a reconstruction branch and an editing branch, both utilizing the same conditional controllers. 
Both branches take random noise as the initial input, and we also compute poses from the source and target videos as inputs to the PoCtr for reconstruction and editing as required.
To further keep the details and textures in the background and foreground, we propose multiple losses to force the consistency between edited and reconstructed features, encapsulated into a regularization term in the score function.


\subsection{Conditional Controller}
\label{sec:controller}
\noindent\textbf{Pose Controller.}
The Pose Controller (PoCtr) modifies the motion in the source according to a series of pose signals. While ControlNet~\cite{zhang2023adding} performs well in conditioned generation, it contains numerous costly attention operations and compromises its robustness when directly generating modified motion from random~\cite{tu2024motioneditor}. 
Therefore, we propose an extremely lightweight module that effectively leverages pose guidance during denoising. PoCtr consists of four convolution blocks, each with two convolution layers. Pose signals are encoded into representations of the same dimension as the initial latents, which are then added to the latents before denoising. This approach is lightweight yet effective compared to conventional signal fusion strategies such as cross-attention and concatenation. We initialize the weights of PoCtr's final projection layer to zero to avoid large influences at the beginning.


\noindent\textbf{Reference Controller.}
To preserve the source's appearance, we introduce the Reference Controller (ReCtr) as a lightweight alternative to the time-consuming DDIM inversion~\cite{song2020denoising} that can cause significant infidelity~\cite{mokady2023null}. Similar to PoCtr, ReCtr includes four convolution blocks for downsampling source features to the same dimension as the latents, with each block containing two convolution layers.
Further, we effectively inject the multi-scale source features into the U-Net by adding to the features of each block in its encoder. Existing comparable designs, \emph{e.g.} MagicAnimate~\cite{xu2023magicanimate} and AnimateAnyone~\cite{hu2023animate}, both leverage the U-Net as the reference controlling module, including numerous complicated attention operations. Additionally, both of them implement concatenation operations for introducing source features to denoising U-Net, increasing the channel dimension during attention calculation, thereby significantly boosting the computation consumption.

To maintain precise consistency and low-level details, we do not use textual prompt-based cross-attention. Instead, source frames are additionally transformed into CLIP embedding using CLIPProjector~\cite{radford2021learning}, and subsequently integrated into U-Net by performing cross-attention with the latents~\cite{hu2023animate}.

\subsection{Training Strategy}
\label{sec:training}
The training process contains two stages, \emph{i.e.}, the image training stage and the video training stage respectively. In the first stage, we utilize individual images for training, in which the motion modules~\cite{guo2023animatediff} are temporarily removed for adapting to the image domain. 
Both the source and target images are randomly extracted from the same video clip. The PoCtr, ReCtr, and denoising U-Net are trainable in this stage.
The weights of two conditional controllers are initialized from Gaussian and the U-Net is based on the pretrained SD1.5 \cite{rombach2022high}. The objective of the first stage is to enable MotionFollower to perform image editing given a source image and a target pose. 

In the second stage, we incorporate motion modules, \emph{e.g.}, temporal layers, into the above-trained MotionFollower for adapting to videos. The temporal layers are trainable while fixing the rest of MotionFollower. 
Following previous practices~\cite{hu2023animate,xu2023magicanimate,zhu2024champ}, the weights of temporal layers are initialized by pretrained AnimateDiff~\cite{guo2023animatediff}. It is noticeable that training temporal layers solely on video clips may deteriorate the model's ability to perform high-quality conditioned editing. To address this issue, we conduct a hybrid training scheme, including image-based editing and video-based editing. 
With a probability of $40\%$, we sample two images as in the first stage for image editing training to maintain single-frame editing fidelity. Otherwise, we randomly select two distinct video clips from the same source video for training, with one acting as the source and the other as the target.
This stage significantly enhances the temporal coherence of MotionFollower.

\subsection{Consistency Guidance via Score Regularization in Inference}
\label{sec:score_guidance}

Although extra signals (target poses and source frames) help control the structure and appearance, the preservation of foreground and background details is not assured due to the lack of supervision in end-to-end training. The reversed diffusion process (\emph{i.e.}, the denoising process) can be derived once the score of the distribution (\emph{i.e.}, the gradient of the log probability density w.r.t. data) is known at each time step \citep{ho2020denoising,song2020score}.
The score function tends to guide the denoising process in a particular direction.
This indicates we can easily add constraints by imposing regularization on the score when estimating it to force the model to denoise the data in our desired direction, thereby solving the intractable inconsistency.

Specifically, we first leverage a lightweight segmentation model (see the Appendix. \ref{sec:person_segmentation}) to obtain the source foreground mask $\bm{M}^{r}$ and predict the edited foreground mask $\bm{M}^{e}$ (\emph{i.e.}, the mask of the source protagonist aligning with the target pose). 
The segmentation model takes a reference image with a pose as input to predict a mask aligning with the given reference and pose. 
$\bm{M}^{r}$ is obtained by sending the source frame and source pose to the segmentation model, while $\bm{M}^{e}$ is predicted by sending the source frame and the target pose.
Then, given the intermediate feature $\bm{F}_{t}^{r},\bm{F}_{t}^{e}$ from the reconstruction and editing branches at the $t$-th step, we formulate our target score function by introducing consistency guidance as follows,
\begin{equation}\small
\label{eq:score_function}
\begin{aligned}
     \nabla_{\bm{z}_{t}^{e}}\log{q}(\bm{z}_{t}^{e},\bm{F}_{t}^{e},\bm{F}_{t}^{r})=\nabla_{\bm{z}_{t}^{e}}\log{q}(\bm{z}_{t}^{e})+\nabla_{\bm{z}_{t}^{e}}\log{q}(\bm{F}_{t}^{e},\bm{F}_{t}^{r}~|~\bm{z}_{t}^{e}),
\end{aligned}
\end{equation}
where $\bm{z}_{t}^{e}$ is the latent in denoising U-Net of the editing branch, and $q(\cdot)$ is the density distribution. 
The first term on the right-hand side is the original score for denoising, and the second term $\nabla_{\bm{z}_{t}^{e}}\log{q}(\bm{F}_{t}^{e},\bm{F}_{t}^{r}~|~\bm{z}_{t}^{e})$ is the consistency regularization, which aims to guide the denoising process in the desired direction, ensuring the generated results to have consistent appearance as the source.

To refine both foreground and background quality in edited results, we design two comprehensive losses $\mathcal{L}_{fg}$ and $\mathcal{L}_{bg}$, whose variables are $\bm{F}_{t}^{e}$ and $\bm{F}_{t}^{r}$, to approximate the score $\nabla_{\bm{z}_{t}^{e}}\log{q}(\bm{F}_{t}^{e},\bm{F}_{t}^{r}~|~\bm{z}_{t}^{e})$. 
Therefore, the score can be converted to $\frac{d(\mathcal{L}_{fg}\odot \bm{M}^{e}+\mathcal{L}_{bg}\odot (1-\bm{M}^{e}))}{d\bm{z}_{t}^{e}}$.
The $fg$ and $bg$ in the $\mathcal{L}_{fg}$ and $\mathcal{L}_{bg}$ refer to protagonists and backgrounds in the edited results, respectively.
Utilizing $\bm{M}^{e}$ to combine $\mathcal{L}_{fg}$ and $\mathcal{L}_{bg}$ aims at performing precise regional updates on the latents concerning foreground and background.
Concretely, $\mathcal{L}_{fg}$ aims to maximize the similarity between the foreground features of the editing and reconstruction branches, enforcing the model to keep the protagonist's appearance:
\begin{equation}\small
\label{eq:foreground_loss}
\begin{aligned}
     \mathcal{L}_{fg}(\bm{F}_{t}^{e},\bm{F}_{t}^{r})=\frac{\alpha_{fg}}{1+2 \cdot \mathtt{Sim}(\mathtt{Pool}(\bm{F}_{t}^{e}\odot\bm{M}^{e}), \mathtt{Pool}(\bm{F}_{t}^{r}\odot\bm{M}^{r}))},
\end{aligned}
\end{equation}
where $\mathtt{Sim}(\cdot)$ and $\mathtt{Pool}(\cdot)$ refer to spatially averaged cosine similarity and mask pooling~\cite{xu2023open}, respectively, and $\alpha_{fg}$ is a hyper-parameter. 
The spatial sizes of $\bm{F}_{t}^{e}$, $\bm{F}_{t}^{r}$, $\bm{M}^{e}$, and $\bm{M}^{r}$ are interpolated into the same size $H\times W$, thus  $\bm{F}^{e},\bm{F}^{r},\bm{M}^{e},\bm{M}^{r} \in {\cal{R}} ^ {T \times N \times H \times W}$. $N$ is the number of frames in the video clip and $T$ is the total number of denoising steps. 

Further, $\mathcal{L}_{bg}$ consists of three loss sub-functions: the background overlapping loss $\mathcal{L}_{over}$, body non-overlapping loss $\mathcal{L}_{body}$, and complementary loss $\mathcal{L}_{com}$. $\mathcal{L}_{over}$ impels the diffusion model to reconstruct overlapping background between the editing and reconstruction features during denoising:
\begin{equation}\small
\label{eq:overlapping_loss}
\begin{aligned}
     \bm{M}^{over}&=(1-\bm{M}^{e}) \odot (1-\bm{M}^{r}), \\
     \mathcal{L}_{over}(\bm{F}_{t}^{e},\bm{F}_{t}^{r})&=\frac{1}{1+2 \cdot \mathtt{Avg}(\mathtt{\sum}_{\substack{k,i,j, \bm{M}^{over}_{t,k,i,j}=1}}\mathtt{Cos}(\bm{F}_{t,k,i,j}^{e}, \bm{F}_{t,k,i,j}^{r}))},
\end{aligned}
\end{equation}
where $\bm{M}^{over}$ refers to the overlapping background regions between the source and the edited videos. $\mathtt{Avg}(\cdot)$ indicates the average operation and $\mathtt{Cos}(\cdot)$ is the cosine similarity.

When ReCtr sends the appearance feature extracted from the source to the editing branch, it potentially introduces source pose bias into the feature distribution. Thus, poses that align with the source rather than the target may lead to the appearance of source poses in areas that should be the background in edited results. 
$\mathcal{L}_{body}$ mitigates pronounced ghosting and blurring effects due to motion misalignment between the source video and the target video:
\begin{equation}\small
\label{eq:body_loss}
\begin{aligned}
     \bm{M}^{body}&=\bm{M}^{r}\odot(1-\bm{M}^{e}), \\
     \mathcal{L}_{body}(\bm{F}_{t}^{e},\bm{F}_{t}^{r})&=\mathtt{Avg}(\mathtt{\sum}_{\substack{k,i,j, \bm{M}^{body}_{t,k,i,j}=1}}\mathtt{Cos}(\bm{F}_{t,k,i,j}^{e}, \bm{F}_{t,k,i,j}^{r})),
\end{aligned}
\end{equation}
where $\bm{M}^{body}$ refers to misaligned motion parts (non-overlapping parts of the protagonist) between the source video and the edited video. 

Finally, $\mathcal{L}_{com}$ ensures that the content of the non-overlapping protagonist's parts closely resembles the background of the entire source video. Specifically, it advances the diffusion model to inpaint non-overlapping protagonist's parts between the source video and the edited video, leveraging background information from the source:
\begin{equation}\small
\label{eq:complement_loss}
\begin{aligned}
     &\mathcal{L}_{com}(\bm{F}_{t}^{e},\bm{F}_{t}^{r})=\frac{1}{1+2 \cdot \mathtt{Sim}(\mathtt{Pool}(\bm{F}_{t}^{e}\odot\bm{M}^{body}), \mathtt{Pool}(\bm{F}_{t}^{r}\odot(1-\bm{M}^{r})))}.
\end{aligned}
\end{equation}
In summary, $\mathcal{L}_{bg}$ simply combines three sub-functions as follows:
\begin{equation}\small
\label{eq:complement_loss}
\begin{aligned}
     &\mathcal{L}_{bg}=\alpha_{over}\mathcal{L}_{over}+\alpha_{body}\mathcal{L}_{body}+\alpha_{com}\mathcal{L}_{com},
\end{aligned}
\end{equation}
where $\alpha_{over}$, $\alpha_{body}$, and $\alpha_{com}$ balance the loss terms.
The consistency guidance $\nabla_{\bm{z}_{t}^{e}}\log{q}(\bm{F}_{t}^{e},\bm{F}_{t}^{r}~|~\bm{z}_{t}^{e})$ finally combines $\mathcal{L}_{fg}$ and $\mathcal{L}_{bg}$:
\begin{equation}\small
\label{eq:guidance}
\begin{aligned}
     &\nabla_{\bm{z}_{t}^{e}}\log{q}(\bm{F}_{t}^{e},\bm{F}_{t}^{r}~|~\bm{z}_{t}^{e})\coloneq\frac{d\mathcal{L}_{fg}}{d\bm{z}_{t}^{e}}\odot \bm{M}^{e}+\frac{d\mathcal{L}_{bg}}{d\bm{z}_{t}^{e}}\odot (1-\bm{M}^{e}).
\end{aligned}
\end{equation}
We exploit our regularization in inference to optimize the latents, rather than the U-Net. 
The sampling process and illustration of losses are in Appendix. \ref{sec:sampling_details}.

\section{Experiments}
\label{sec:experiments}

\subsection{Implementation Details}
\label{sec:implementation_details}

To demonstrate the superiority of our model in modifying various motions, we collect 3K videos (60-90 seconds long) from the internet to train our model, in which all video frames are cropped to a resolution of $512\times512$. 
We utilize DWPose~\cite{yang2023effective} and our proposed lightweight person segmentation method to extract poses and masks. The details of our proposed person segmentation method are depicted in the Appendix. \ref{sec:person_segmentation}. We evaluate our model on 100 unseen in-the-wild cases from YouTube and the TikTok dataset~\cite{wang2023disco}. The training process is conducted on 4 NVIDIA A100 GPUs. The first and second training stages are both trained for 100,000 steps with a batch size of 64 and 4 respectively. The learning rates of the two stages are set to $1e-5$. In inference, we first align the target pose sequence with the source by introducing the pose alignment algorithm~\cite{tu2024motioneditor}. 
MotionFollower requires 50 seconds to perform motion editing of a video with 24 frames on a single NVIDIA A100.

\subsection{Motion Editing Results}
We show our motion editing results in Fig. \ref{fig:general_case}, and more results are illustrated in Fig. \ref{fig:editing_result}-\ref{fig:supp-camera} of the Appendix. \ref{sec:additional_results}.
MotionFollower can manipulate the motions of various protagonists in videos featuring large camera movements and complicated appearances, including backgrounds, initial poses, and the protagonist's clothing. 
We can see that our method can handle various motion editing while maintaining appearance and temporal consistency with the source videos. 

\begin{figure}[t!]
\begin{center}
\includegraphics[width=1\linewidth]{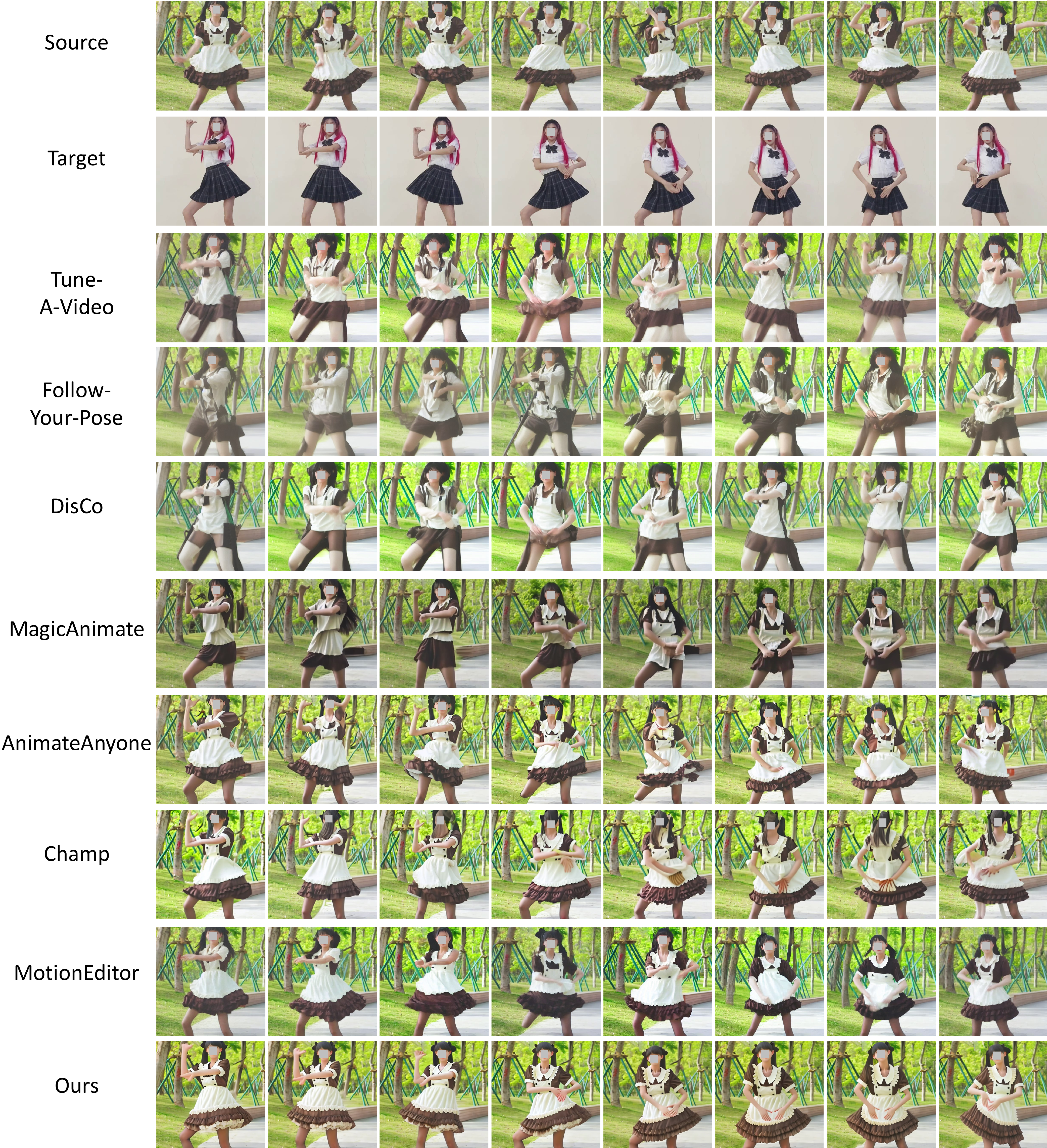}
\end{center}
\vspace{-0.3cm}
   \caption{\small Qualitative comparison between our MotionFollower and other state-of-the-art models. 
   Our method exhibits accurate motion editing and appearance preservation.
   }
\label{fig:comparison_result}
\vspace{-0.3cm}
\end{figure}

\subsection{Comparison with State-of-the-art Methods}
\textbf{Competitors.} We compare with recent pose-driven models. The competitors include MRAA~\cite{siarohin2021motion}, Tune-A-Video~\cite{wu2023tune}, Follow-Your-Pose~\cite{ma2023follow}, Disco~\cite{wang2023disco}, MagicAnimate~\cite{xu2023magicanimate}, AnimateAnyone~\cite{hu2023animate}, Champ~\cite{zhu2024champ}, and MotionEditor~\cite{tu2024motioneditor}. 
Particularly, Tune-A-Video is equipped with ControlNet~\cite{zhang2023adding} to enable controllable editing.
It is worth noting that human motion transfer models (MRAA, Disco, MagicAnimate, AnimateAnyone, Champ) have been adapted to support video motion editing by sending all source frames to their reference networks instead of just the original single source image.


\textbf{Qualitative results.} We conduct a qualitative comparison between our MotionFollower and several competitors in Fig. \ref{fig:comparison_result}. By analyzing the results, we can gain the following observations: (1) All competitors can manipulate the motion of the given video to some extent. However, their edited results exhibit significant ghosting issues, where frames of the protagonist's heads and legs overlap, creating strong artifacts. 
(2) Tune-A-Video~\cite{wu2023tune} and Follow-Your-Pose~\cite{ma2023follow} encounter severe overlapping effect and fail to preserve the protagonist's appearance. The plausible reason is that they do not interact with the source intermediate features. (3) In terms of the human motion transfer methods (Disco~\cite{wang2023disco}, MagicAnimate~\cite{xu2023magicanimate}, AnimateAnyone~\cite{hu2023animate}, Champ~\cite{zhu2024champ}), they fail to perform accurate motion manipulation and reconstruct the protagonist's clothes with high fidelity.
The results indicate that existing human motion transfer models encounter challenges when performing motion editing on videos containing complex information. (4) Finally, our MotionFollower excels at precisely modifying motion in videos while maintaining the integrity of the source background and appearance, showcasing its significant superiority compared to previous methods.

\textbf{Quantitative results.} Following the previously established evaluation metrics, we take the single frame quality (L1 error, SSIM~\cite{wang2004image}, LPIPS~\cite{zhang2018unreasonable}, PSNR~\cite{hore2010image}, and FID~\cite{heusel2017gans}) and video fidelity (FID-FVD~\cite{balaji2019conditional}, FVD~\cite{unterthiner2018towards}) into consideration. We partitioned a test video into two segments: the first segment serves as the input, while the second segment serves as the target. We conduct the validation on 100 collected in-the-wild videos, comprising complex backgrounds and protagonists' appearances. 
Table \ref{table:quantitative_comparisons} depicts the comparison of quantitative results between our method and other competitors. We can observe that our MotionFollower outperforms all competitors in terms of single-frame quality and video fidelity at a cost-efficient rate. Our model outperforms the best competitor MotionEditor by achieving 8.15 FID-VID and 119.39 FVD with an approximately 80\% reduction in GPU consumption, demonstrating the superiority of our model among other competitors. 
With a single NVIDIA A100, MotionEditor can only edit 16 frames at a time, while our MotionFollower can edit 56 frames. We also show more subjective evaluation results (human evaluation) in the Appendix. \ref{sec:user_study}.

\begin{table}[t!]\small
\caption{\small Quantitative comparison on 100 wild cases. Mem refers to GPU memory when manipulating 8 frames.
}
\vspace{-0.15in}
\begin{center}
\renewcommand\arraystretch{1.1}
\scalebox{0.85}{
\begin{tabular}{lcccccccc}
\toprule
Model            & L1~$\downarrow$               & PSNR~$\uparrow$           & SSIM~$\uparrow$         & LPIPS~$\downarrow$         & FID~$\downarrow$           & FID-VID~$\downarrow$       & FVD~$\downarrow$            & Mem~$\downarrow$         \\ \midrule
MRAA~\cite{siarohin2021motion}             & 7.83E-4          & 8.03           & 0.35          & 0.67          & 120.51         & 100.12         & 783.72          & \textbf{6.4G}          \\
Tune-A-Video~\cite{wu2023tune}     & 3.88E-4          & 12.01          & 0.42          & 0.51          & 66.37          & 84.58          & 613.44          & 31.7G         \\
Follow-Your-Pose~\cite{ma2023follow} & 4.55E-4          & 11.63          & 0.37          & 0.55          & 73.12          & 87.65          & 672.27          & 7.2G          \\
Disco~\cite{wang2023disco}            & 2.92E-4          & 9.24           & 0.38          & 0.47          & 92.97          & 83.69          & 635.08          & 18.3G         \\
MagicAnimate~\cite{xu2023magicanimate}     & 1.09E-4          & 16.22          & 0.62          & 0.35          & 33.04          & 26.59          & 477.65          & 21.8G         \\
AnimateAnyone~\cite{hu2023animate}    & 9.45E-5          & 16.18          & 0.65          & 0.32          & 35.81          & 28.31          & 515.57          & 16.1G         \\
Champ~\cite{zhu2024champ}            & 9.94E-5          & 16.12          & 0.58          & 0.36          & 36.55          & 25.89          & 452.65          & 17.4G         \\
MotionEditor~\cite{tu2024motioneditor}     & 9.13E-5          & 17.34          & 0.68          & 0.34          & 31.98          & 20.57          & 395.43          & 42.6G         \\ \midrule
MotionFollower   & \textbf{6.31E-5} & \textbf{20.85} & \textbf{0.75} & \textbf{0.22} & \textbf{26.30} & \textbf{12.42} & \textbf{276.04} & 9.8G \\ \bottomrule
\end{tabular}
}
\end{center}
\label{table:quantitative_comparisons}
\vspace{-0.25in}
\end{table}

\subsection{Discussion}

\noindent\textbf{Long Video Synthesis.} We conduct a qualitative comparison between our proposed MotionFollower and the strongest competitor MotionEditor in Fig. \ref{fig:long_video} of the Appendix. \ref{sec:additional_results}. The source video comprises 600 frames containing intricate protagonist's clothes, complex initial poses, and dynamic camera movements. The results depict that the edited frames of MotionEditor encounter numerous blurry noises and human body distortion. A plausible explanation is that camera movements introduce dynamism to objects and backgrounds, making it challenging for conventional diffusion models to capture semantic information accurately. By contrast, due to our specialized consistency guidance, our model can effectively perform precise motion editing on long videos with dynamic camera movements while preserving the original information.

\noindent\textbf{Camera Movement.} Our proposed consistency guidance can force the diffusion model to concentrate on the semantic details of the source, thereby promoting the model to preserve the original background information, including camera movements. Fig. \ref{fig:camera_movement} in the Appendix. \ref{sec:additional_results} demonstrates the comparison results. Large camera movements can deteriorate the editing capability of all competitors, as they struggle to maintain the semantic details of the dynamic background, thereby exacerbating the occurrence of blurring noise. In contrast, our proposed MotionFollower can effectively perform video motion editing while preserving complex camera movements and other background details.


\begin{table}[t!]\small
\caption{\small Quantitative comparison on the Human Motion Transfer task on TikTok benchmark~\cite{jafarian2021learning}.
}
\vspace{-0.15in}
\begin{center}
\renewcommand\arraystretch{1.1}
\scalebox{0.8}{
\begin{tabular}{lcccccc}
\toprule
Model          & L1~$\downarrow$               & PSNR~$\uparrow$           & SSIM~$\uparrow$           & LPIPS~$\downarrow$          & FID-VID~$\downarrow$        & FVD~$\downarrow$             \\ \midrule
MRAA~\cite{siarohin2021motion}           & 3.21E-4          & 29.39          & 0.672          & 0.296          & 54.47          & 284.82          \\
Disco~\cite{wang2023disco}          & 3.78E-4          & 29.03          & 0.668          & 0.292          & 59.90          & 292.80          \\
MagicAnimate~\cite{xu2023magicanimate}   & 3.13E-4          & 29.16          & 0.714          & 0.239          & \textbf{21.75}         & 179.07          \\
AnimateAnyone~\cite{hu2023animate}  & -                & 29.56          & 0.718          & 0.285          & -              & 171.9           \\
Champ~\cite{zhu2024champ}          & 3.02E-4          & \textbf{29.84}          & 0.773          & 0.235          & 26.14          & 170.20          \\ \midrule
MotionFollower & \textbf{2.89E-4} & {29.25} & \textbf{0.793} & \textbf{0.230} & 22.36 & \textbf{159.88} \\ \bottomrule
\end{tabular}
}
\end{center}
\label{table:animation_comparisons}
\vspace{-0.10in}
\end{table}

\noindent\textbf{Human Motion Transfer.} Video motion editing conducts motion transfer directly on the source video over the temporal dimension, which involves multiple initial poses, camera movements, and dynamic backgrounds. 
This task is typically more challenging than human motion transfer, which only focuses on animating a single image with a simple initial pose and a static background while neglecting dynamic information.
To demonstrate that MotionFollower also excels over other methods in the less challenging motion transfer tasks, we convert the workflow of motion editing into the workflow of human motion transfer for a comparative analysis of motion manipulation capabilities. 
We follow the evaluation protocol as DisCo~\cite{wang2023disco}, training and testing our model on TikTok~\cite{jafarian2021learning} for fair comparison. The quantitative comparisons are presented in Table \ref{table:animation_comparisons}. Our method achieves the best results in most metrics, indicating that our model can effectively handle conditioned image animation due to the promising ability of motion manipulation.

We also illustrate a qualitative comparison between our MotionFollower and other state-of-the-art models in terms of human motion transfer, as presented in Fig. \ref{fig:animation} and \ref{fig:animation-plus} of the Appendix. \ref{sec:additional_results}. It is noticeable that our model can perform gorgeous image animation, while the results of other competitors suffer from body distortion and a dramatic degree of ghosting from overlapping hands. The results indicate that our model is also capable of human motion transfer.

\begin{table}[t!]\small
\caption{\small Ablations on key components. \textit{+RNet} and \textit{+CNet} refers to inserting the Reference Net (RNet) from MagicAnimate~\cite{xu2023magicanimate} and ControlNet (CNet)~\cite{zhang2023adding} into our model, respectively.
}
\vspace{-0.15in}
\begin{center}
\renewcommand\arraystretch{1.1}
\scalebox{0.8}{
\begin{tabular}{lcccccccc}
\toprule
Model                & L1~$\downarrow$               & PSNR~$\uparrow$           & SSIM~$\uparrow$          & LPIPS~$\downarrow$         & FID~$\downarrow$            & FID-VID~$\downarrow$        & FVD~$\downarrow$             & Mem~$\downarrow$           \\ \midrule
\textit{w/o} ReCtr            & 2.78E-4          & 9.03          & 0.42          & 0.45          & 64.82          & 47.28          & 545.39          & 8.3G          \\
(\textit{w/o} ReCtr)\textit{+RNet}       & 7.33E-5          & 19.82          & 0.72          & 0.25          & 26.35          & 13.12          & 288.57          & 17.8G         \\
(\textit{w/o} PoCtr)\textit{+CNet} & 8.27E-5          & 18.13          & 0.65          & 0.28          & 30.57          & 23.75          & 381.52          & 15.2G         \\
\textit{w/o} score guidance   & 9.78E-5          & 16.50          & 0.61          & 0.36          & 35.91          & 28.10          & 437.69          & \textbf{7.1G}          \\ \midrule
Ours                 & \textbf{6.31E-5} & \textbf{20.85} & \textbf{0.75} & \textbf{0.22} & \textbf{26.30} & \textbf{12.42} & \textbf{276.04} & 9.8G \\ \bottomrule
\end{tabular}
}
\end{center}
\label{table:ablation}
\vspace{-0.4cm}
\end{table}

\subsection{Ablation Study}
\begin{figure}[t!]
\begin{center}
\includegraphics[width=0.95\linewidth, trim=0.1cm 0cm 0cm 0cm, clip]{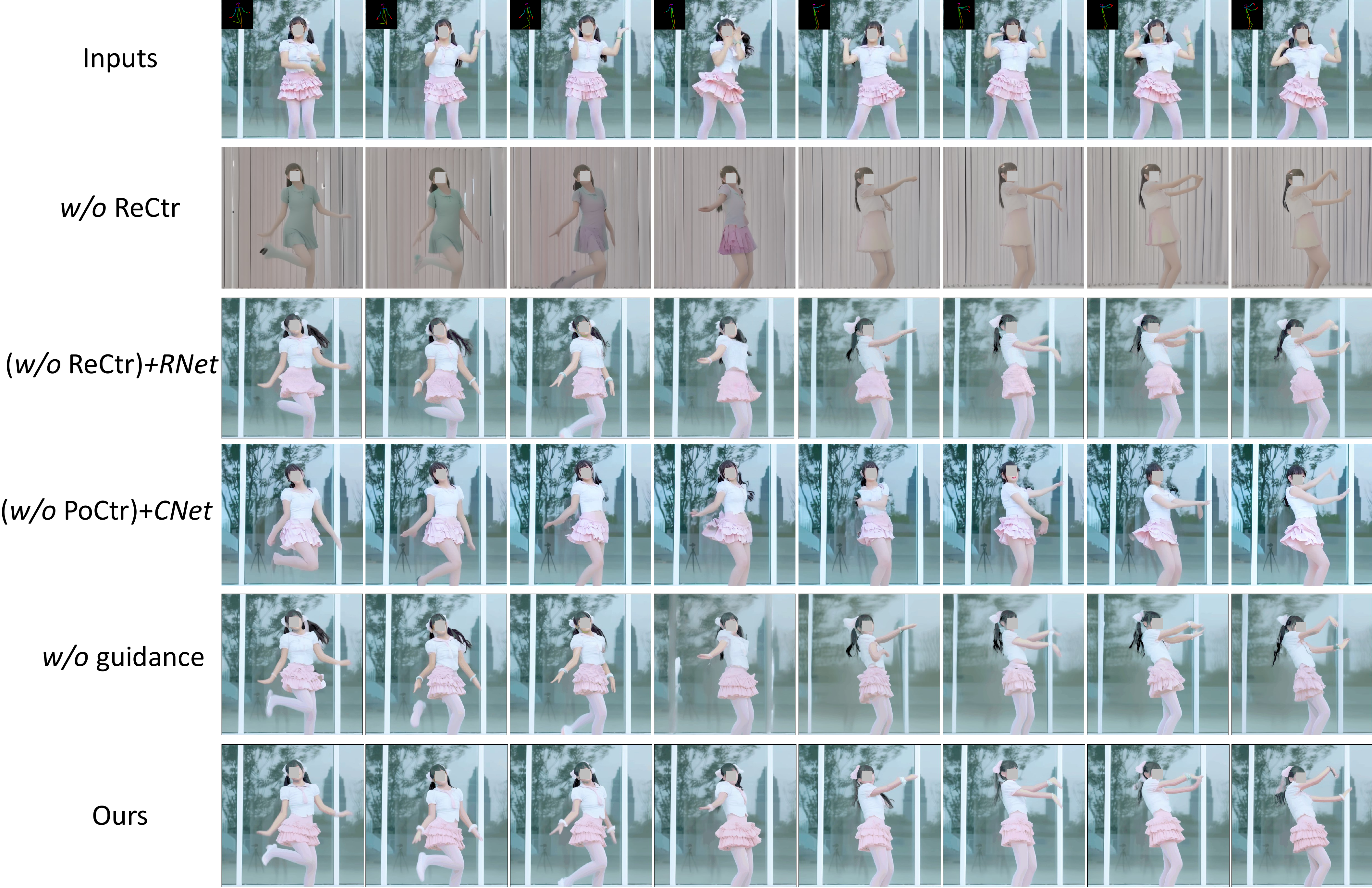}
\end{center}
\vspace{-0.5cm}
   \caption{\small Example illustration of ablation study on core components of the proposed MotionFollower.
   }
\label{fig:ablation}
\vspace{-0.2in}
\end{figure}

We conduct an ablation study to validate the significance of the core components in MotionFollower. The quantitative and qualitative results are depicted in Table \ref{table:ablation} and Fig. \ref{fig:ablation}. 
We replace our proposed ReCtr and PoCtr with Reference Net (RNet) in MagicAnimate~\cite{xu2023magicanimate} and ControlNet (CNet)~\cite{zhang2023adding}, respectively, to demonstrate their effectiveness and efficiency. 
We can observe that \textit{w/o} ReCtr has a significantly negative impact on the editing performance, as the model solely relies on cross-attention with the source embedding, which fails to reconstruct the original appearance.
(\textit{w/o} ReCtr)\textit{+RNet} leads to slightly inferior performance compared to our model. However, the UNet-based Reference Net consumes roughly twice the GPU memory as our proposed model. 
In the third row of Fig. \ref{fig:ablation}, the protagonist's clothes are considerably blurry compared to our model, indicating that our ReCtr can retain more appearance details due to pure convolution operations.
Similarly, (\textit{w/o} PoCtr)\textit{+CNet} can also exacerbate the strain on GPU memory consumption, as ControlNet consists of multiple costly attention operations. We can also observe that replacing our PoCtr with the relatively complicated ControlNet can degrade the editing effectiveness. The plausible reason is that complex modeling on poses may potentially introduce blurry noise to the diffusion model, destroying the distribution relationship between the reference net and the denoising U-Net. The results without score guidance suffer from a blurry background and disappearance of some semantic details, validating that our score guidance can drastically retain the background and content consistency. We also conduct an ablation study in terms of different loss functions, as shown in the Appendix. \ref{sec:additional_ablation}.


\section{Related Work}

\noindent\textbf{Diffusion for Video Editing and Generation.} The success of diffusion-based image generation has inspired numerous studies in diffusion-based video generation~\cite{esser2023structure,ho2022video,hong2022cogvideo,khachatryan2023text2video,singer2022make}. Most current T2V models inflate pretrained T2I diffusion models by inserting additional temporal layers. AnimateDiff~\cite{guo2023animatediff} proposes a motion module that can be directly plugged into other personalized diffusion models. Meanwhile, pose-guided video synthesis is particularly prevalent. Follow-Your-Pose~\cite{ma2023follow} and ControlVideo~\cite{zhang2023controlvideo} insert an adapter into the T2I diffusion model for modeling the pose sequence. 
However, the video generation models mainly focus on diversity instead of preservation. 

Most video editing approaches can be divided into two categories, namely the mutual attention-based method and the layered neural atlas-based method. Tune-A-Video~\cite{wu2023tune}, Text2Video-Zero~\cite{khachatryan2023text2video}, FateZero~\cite{qi2023fatezero}, CCEdit~\cite{feng2023ccedit} belong to the first class, while Text2LIVE~\cite{bar2022text2live}, StableVideo~\cite{chai2023stablevideo}, and VidEdit~\cite{couairon2023videdit} belong to the latter. 
However, the above methods primarily focus on low-level attribute editing and are unable to edit complex motions. Regarding video motion editing, MotionEditor~\cite{tu2024motioneditor} stands as the most advanced diffusion model equipped for this task, achieved through applying a motion adapter and a complex attention injection, escalating its computational demands.
Further, its attention injection struggles with certain videos with complex backgrounds or camera movements.

\noindent\textbf{Human Motion Transfer.} This task aims to animate images based on one given image. 
Previous GAN-based models~\cite{liu2019liquid,siarohin2019first,siarohin2021motion,liu2023human,huang2021few} struggle with complex backgrounds and poses. Recently, some studies have applied diffusion models to this field.
DisCo~\cite{wang2023disco} uses masks to decouple the objects for modeling. MagicAniamte~\cite{xu2023magicanimate} leverages ControlNet and an additional U-Net for modeling poses and appearances independently. AnimateAnyone~\cite{hu2023animate} combines attention features from the reference net with those of a denoising U-Net for joint self-attention. Champ~\cite{zhu2024champ} introduces SMPL for enhancing controlling. However, these methods are unable to handle complex motions while preserving the original camera movements and appearances. 
They also require the input image to have a relatively clear initial pose, which may not be realistic in practice. 
By contrast, MotionFollower can synthesize consistent videos in complex motion while preserving original dynamic information.

\section{Conclusion}
In this paper, we propose MotionFollower for addressing the video motion editing challenge, which remains relatively unexplored and is considered as high-level video editing compared to conventional video attribute editing. To enhance the signal-controlling capability, we propose two lightweight modules ReCtr and PoCtr (one for source frames and the other for poses), which purely consist of convolution operations. We further propose a score guidance principle, empowering the diffusion model to accurately reconstruct both the backgrounds and the details of the protagonist's appearance. 
Experiments demonstrate that our MotionFollower has the superiority of motion manipulation while preserving original details.

{\small
\bibliographystyle{abbrv}

}

\newpage
\appendix
\section{Appendix}
\subsection{Person Segmentation}
\label{sec:person_segmentation}

\begin{figure}[h]
\begin{center}
\includegraphics[width=0.8\linewidth]{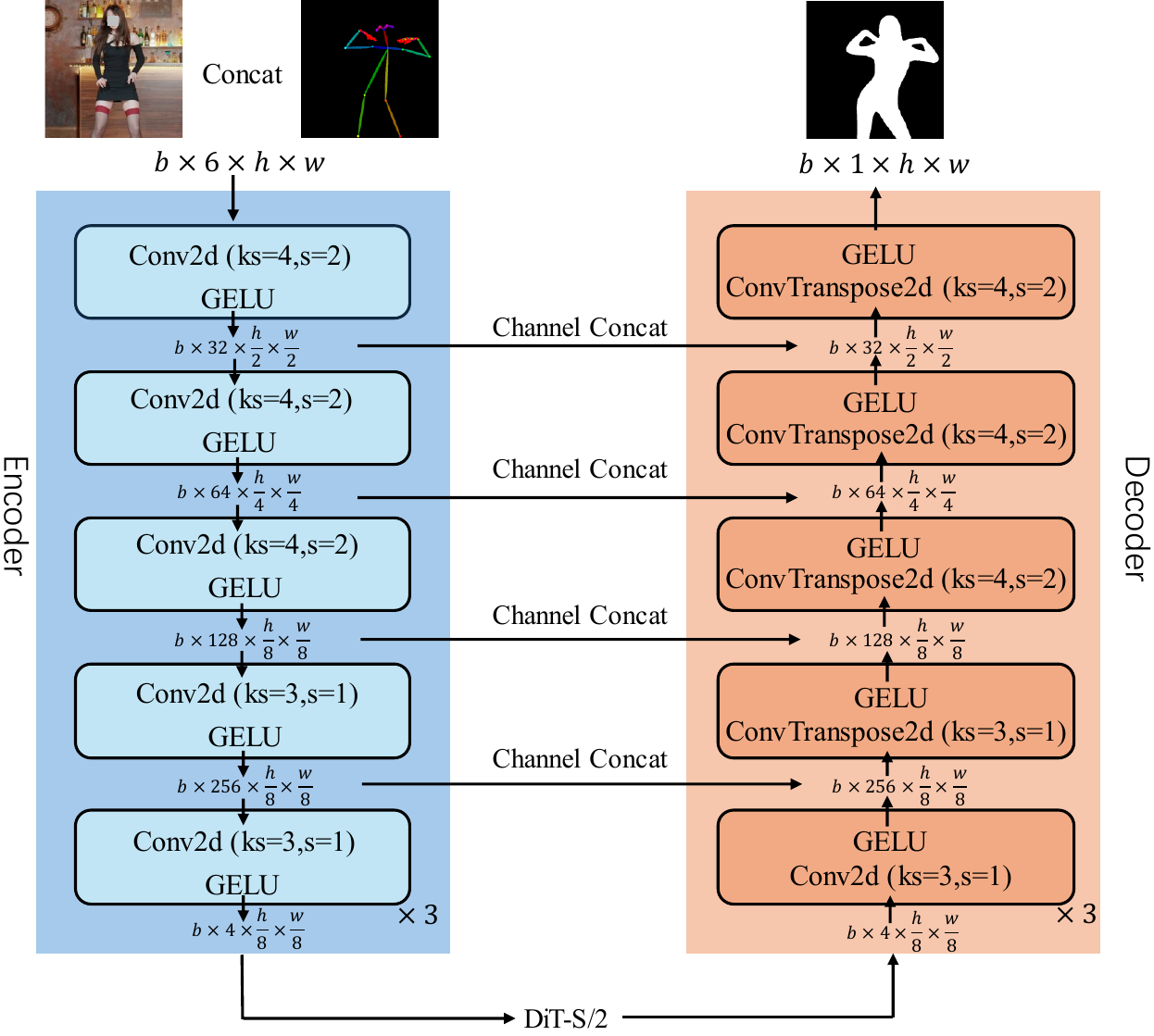}
\end{center}
\vspace{-0.1cm}
   \caption{\small The overview of our proposed person segmentation model.
   }
\label{fig:segmodel}
\vspace{-0.0cm}
\end{figure}

To gain the protagonist mask from the given condition, we propose a lightweight person segmentation model based on DiT~\cite{peebles2023scalable} architecture, where the parameter number of our segmentation model is only 23.5MB compared with SAM-B~\cite{kirillov2023segment} (91MB).
SAM-B requires 40 seconds to extract masks of 24 frames, while our segmentation model only requires 6 seconds to accomplish the same job. 
Given a single image and a target pose, our segmentation model can predict the protagonist mask aligning with the target pose. Concretely, we concatenate the single image with the target pose in the channel dimension as the input of our model. We leverage several convolution blocks to extract different levels of representations, and the resulting features are sent to DiT for further modeling. 
The output of DiT is sent to the decoder to obtain the predicted mask, which is thresholded to obtain a binary mask.
It is worth noting that the intermediate features from the first four convolution modules are concatenated with the outputs of the corresponding convolution modules in the decoder, which can contribute to enhancing controllable modeling.

In terms of training, we train our person segmentation model $\bm{Seg}(\cdot)$ using a 1000 video subset of our collected dataset. We split the entire video into two clips, where the first clip serves as the source and the second clip serves as the target. We obtain the poses and the masks of the protagonist from the target by employing DWPose~\cite{yang2023effective} and SAM~\cite{kirillov2023segment} on the target frames. The source frame $\bm{F}_{sr}$ and the extracted pose $\bm{P}_{tg}$ from the target serve as the inputs of our person segmentation model during training, and the extracted mask $\bm{M}_{tg}$ from the target serves as ground truth. We implement the MSE loss to train our model, which can be described as:
\begin{equation}\small
\label{eq:segmentation}
\begin{aligned}
     \bm{M}^{pred}&=\bm{Seg}(\bm{F}_{sr},\bm{P}_{tg}), \\
     \mathcal{L}_{mse}(\bm{M}^{pred},\bm{M}_{tg})&=\left \| \bm{M}^{pred}-\bm{M}_{tg} \right \|_{2}^{2}.
\end{aligned}
\end{equation}
It is worth noting that all components of our person segmentation model are trainable. 

\subsection{Preliminaries}
Diffusion models~\cite{song2020denoising,song2020score,ho2020denoising} have shown gorgeous results for high-quality image synthesis. They are based on thermodynamics, consisting of a forward diffusion process and a reverse denoising process. During the forward process, models appended to a constant noise schedule $\bm{\alpha}_{t}$ add random noise to the source sample $\bm{x}_{0}$ at time step $t$ for obtaining a noise sample $\bm{x}_{t}$:
\begin{equation}\small
\label{eq:diffusion_forward}
\begin{aligned}
     &\bm{q}(\bm{x}_{1:T}) = \bm{q}(\bm{x}_{0})\prod_{t=1}^{T}\bm{q}(\bm{x}_{t}|\bm{x}_{t-1}), \\
     &\bm{q}(\bm{x}_{t}|\bm{x}_{t-1}) = \mathcal{N}(\bm{x}_{t}; \sqrt{\alpha_{t}}\bm{x}_{t-1},(1-\alpha_{t})\mathbf{I}).
\end{aligned}
\end{equation}
The source sample $\bm{x}_{0}$ is ultimately inverted into Gaussian noise $\bm{x}_{T}\sim \mathcal{N}(\mathbf{0}, \mathbf{1})$ after $T$ forward steps. 
The reverse process recovers $\bm{x}_{0}$ from $\bm{x}_{T}$ by some denoising steps. The denoising network $\bm{\varepsilon}_{\theta}(\bm{x}_{t},t)$ tends to predict the noise $\bm{\varepsilon}$ conditioned on the current sample $\bm{x}_{t}$ and time step $t$ by training with a simplified mean squared error:
\begin{equation}\small
\label{eq:diffusion_loss}
\begin{aligned}
     \mathcal{L}_{simple} = \mathbb{E}_{\bm{x}_{0},\bm{\varepsilon},t}(\left \| \bm{\varepsilon} -\bm{\varepsilon}_{\theta}(\bm{x}_{t}, t)  \right \|^{2}).
\end{aligned}
\end{equation}
Further, diffusion models can be regarded as continuous models~\cite{song2020score}. According to Langevin dynamics~\cite{song2020improved}, the continuous denoising process can be depicted as the score function $\nabla_{\bm{x}_{t}}\log{q}(\bm{x}_{t})$, sampling from the Gaussian noise. In terms of an additional condition $\bm{c}$, the score function can be described as $\nabla_{\bm{x}_{t}}\log{q}(\bm{x}_{t}, \bm{c})$, supporting conditional denoising.

\subsection{Implementation of Consistency Guidance}
\label{sec:sampling_details}

According to previous works~\cite{song2020score, dhariwal2021diffusion}, when the diffusion model $\bm{\varepsilon}_{\theta}(\bm{z}_{t}^{e})$ tends to predict the noise added to the original frames, it can be converted to the form of the score function, which can be depicted as:
\begin{equation}\small
\label{eq:converted_socre}
\begin{aligned}
     \nabla_{\bm{z}_{t}^{e}}\log{q}(\bm{z}_{t}^{e})=-\frac{1}{\sqrt{1-\bm{\bar{\alpha}}_{t}}}\bm{\varepsilon}_{\theta}(\bm{z}_{t}^{e}). 
\end{aligned}
\end{equation}
We implement the additional conditions $\bm{F}_{t}^{e}$ and $\bm{F}_{t}^{r}$ to the score function, which can be described as:
\begin{equation}\small
\label{eq:condition_socre}
\begin{aligned}
     \nabla_{\bm{z}_{t}^{e}}\log{q}(\bm{z}_{t}^{e},\bm{F}_{t}^{e},\bm{F}_{t}^{r})
     &=\nabla_{\bm{z}_{t}^{e}}\log{q}(\bm{z}_{t}^{e})+\nabla_{\bm{z}_{t}^{e}}\log{q}(\bm{F}_{t}^{e},\bm{F}_{t}^{r}|\bm{z}_{t}^{e}) \\
     &=-\frac{1}{\sqrt{1-\bm{\bar{\alpha}}_{t}}}\bm{\varepsilon}_{\theta}(\bm{z}_{t}^{e})+\nabla_{\bm{z}_{t}^{e}}\log{q}(\bm{F}_{t}^{e},\bm{F}_{t}^{r}|\bm{z}_{t}^{e}).
\end{aligned}
\end{equation}
We can ultimately obtain a new noise prediction model $\bm{\hat{\varepsilon}}_{\theta}(\bm{z}_{t}^{e}, \bm{F}_{t}^{e}, \bm{F}_{t}^{r})$ for the joint distribution, which can be described as:
\begin{equation}\small
\label{eq:joint_socre}
\begin{aligned}
     \bm{\hat{\varepsilon}}_{\theta}(\bm{z}_{t}^{e}, \bm{F}_{t}^{e}, \bm{F}_{t}^{r})=\bm{\varepsilon}_{\theta}(\bm{z}_{t}^{e})-\sqrt{1-\bm{\bar{\alpha}}_{t}}\nabla_{\bm{z}_{t}^{e}}\log{q}(\bm{F}_{t}^{e},\bm{F}_{t}^{r}|\bm{z}_{t}^{e}).
\end{aligned}
\end{equation}
Therefore, the sampling process equipped with our proposed consistency guidance can be depicted in Algorithm \ref{alg:sampling}. $\mathtt{guidance}(\cdot)$ and $\mathtt{decoder}(\cdot)$ indicate our proposed consistency guidance and latent diffusion decoder. 
The hyper-parameters $\alpha_{fg}, \alpha_{over}$, $\alpha_{body}$, $\alpha_{com}$ are set to $4.0$, $6.0$, $2.4$, $1.2$. 

\begin{figure}[h]
\begin{center}
\includegraphics[width=0.95\linewidth]{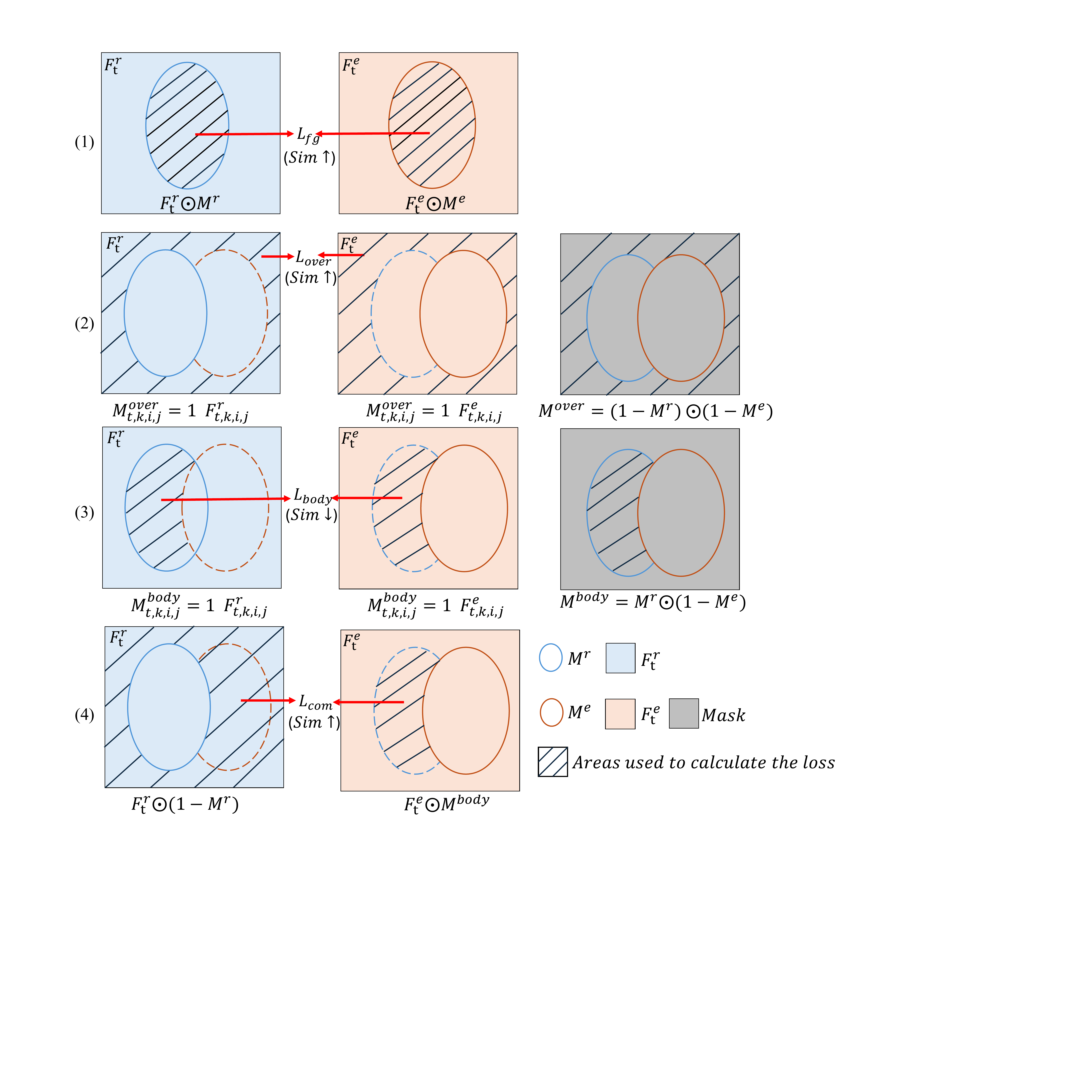}
\end{center}
\vspace{-0.1cm}
   \caption{\small The illustration of our proposed loss functions.
   }
\label{fig:illustrate_loss}
\vspace{-0.0cm}
\end{figure}

Furthermore, we illustrate our proposed 4 loss functions in Figure \ref{fig:illustrate_loss}. The areas with black stripes are used to calculate the loss. $Sim \uparrow$ indicates that the corresponding loss encourages the two areas to have higher similarity and thus a similar appearance, while $Sim \downarrow$ means that the loss aims to push the feature of these two areas away to avoid ghosting artifacts.

\begin{algorithm}[t!]
\caption{Sampling process equipped with our proposed consistency guidance}
\label{alg:sampling}
\begin{algorithmic}
\small 
\State \textbf{Input:} Source video $\bm{V}_{sr}$, Source Mask $\bm{M}_{r}$, Source Pose $\bm{P}_{sr}$; Target Pose $\bm{P}_{tg}$, Predicted Mask $\bm{M}_{e}$ 
 \State $\bm{z}_{T}^{e}\gets$ sample from $\mathcal{N}(\mathbf{0}, \mathbf{I})$
 \State \textbf{for all} $t$ from $T$ to $1$ \textbf{do}
 \State \hspace{1em} $\bm{\varepsilon},\bm{F}_{t}^{e},\bm{F}_{t}^{r}\gets\bm{\varepsilon}_{\theta}(\bm{z}_{t}^{e},\bm{V}_{sr},\bm{P}_{sr},\bm{P}_{tg})$
 \State \hspace{1em} $\nabla_{\bm{z}_{t}^{e}}\log{q}(\bm{F}_{t}^{e},\bm{F}_{t}^{r}|\bm{z}_{t}^{e})\gets\mathtt{guidance}(\bm{z}_{t}^{e},\bm{F}_{t}^{e},\bm{F}_{t}^{r},\bm{M}_{r},\bm{M}_{e})$
 \State \hspace{1em} $\bm{\hat{\varepsilon}}\gets\bm{\varepsilon}-\sqrt{1-\bm{\bar{\alpha}}_{t}}\nabla_{\bm{z}_{t}^{e}}\log{q}(\bm{F}_{t}^{e},\bm{F}_{t}^{r}|\bm{z}_{t}^{e})$
 \State \hspace{1em} $\bm{z}_{t-1}^{e}\gets\sqrt{\bm{\bar{\alpha}}_{t-1}}(\frac{\bm{z}_{t}^{e}-\sqrt{1-\bm{\bar{\alpha}}_{t}}\bm{\hat{\varepsilon}}}{\sqrt{\bm{\bar{\alpha}}_{t}}})+\sqrt{1-\bm{\bar{\alpha}}_{t-1}}\bm{\hat{\varepsilon}}$
 \State \textbf{end for}
 \State $\bm{x}_{0}\gets\mathtt{decoder}(\bm{z}_{0}^{e})$
 \State \textbf{return} $\bm{x}_{0}$  
\end{algorithmic}
\end{algorithm}

\subsection{User Study}
\label{sec:user_study}

We conduct a user study (human evaluation) to validate the human preference between MotionFollower and other competitors. For each case, participants are first presented with the source video and target video. Then we present two motion-edited videos in random order; one is edited by our MotionFollower and the other is generated by a competitor. Participants are asked to answer the questions: ``which one has better motion alignment with the target'', ``which one has better protagonist's appearance alignment with the source'', and ``which one has better background alignment with the source.'' The cases in user study include 100 videos. Participants are mainly university students and faculties. Table \ref{table:user_study} shows the superiority of our model in terms of subjective evaluation.

\subsection{Additional Results}
\label{sec:additional_results}

\begin{figure*}[h]
\begin{center}
\includegraphics[width=0.98\linewidth]{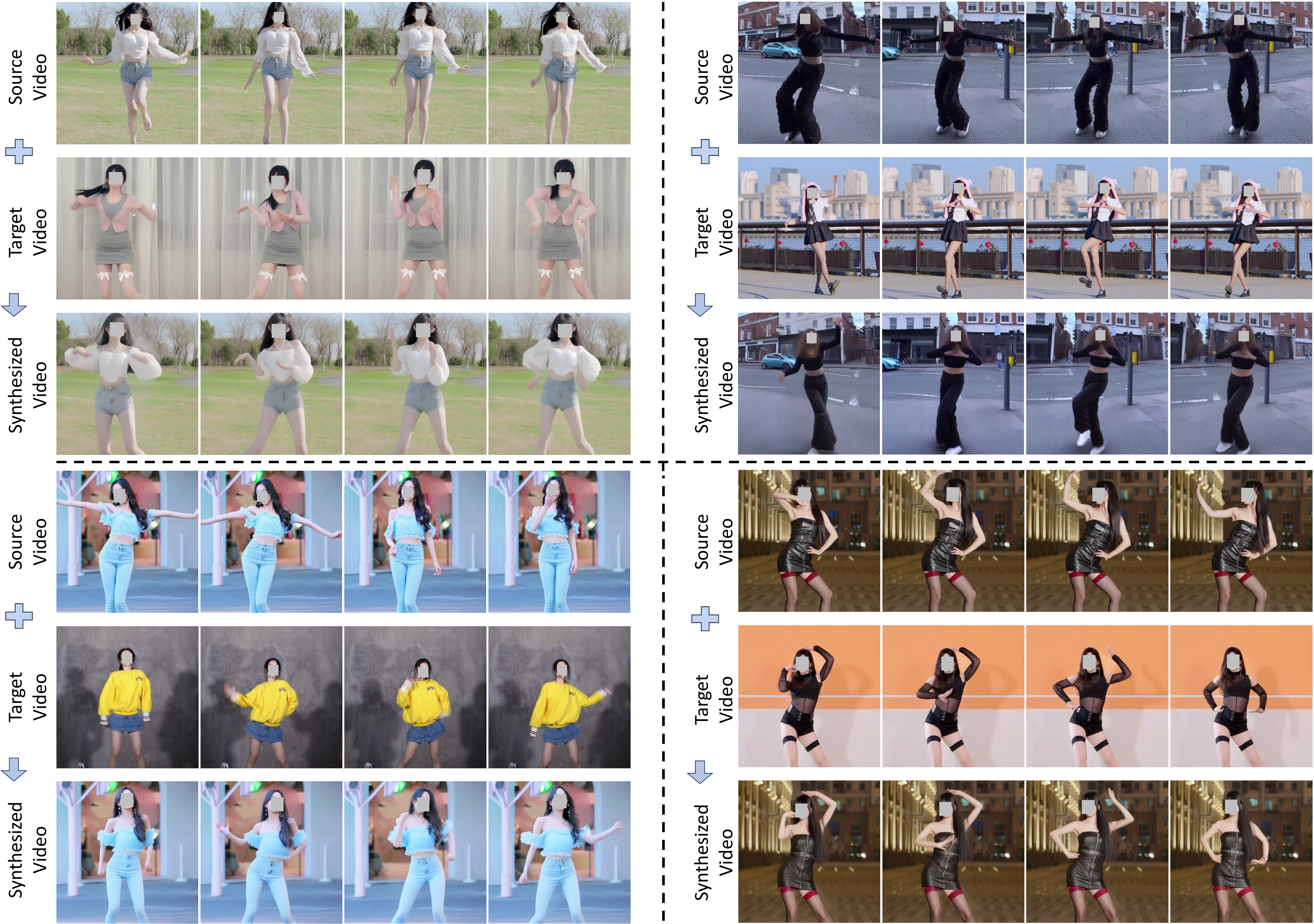}
\end{center}
\vspace{-0.0cm}
   \caption{\small Video motion editing results of our MotionFollower.}
\label{fig:editing_result}
\vspace{3cm}
\end{figure*}

\begin{figure}[H]
\begin{center}
\includegraphics[width=1\linewidth]{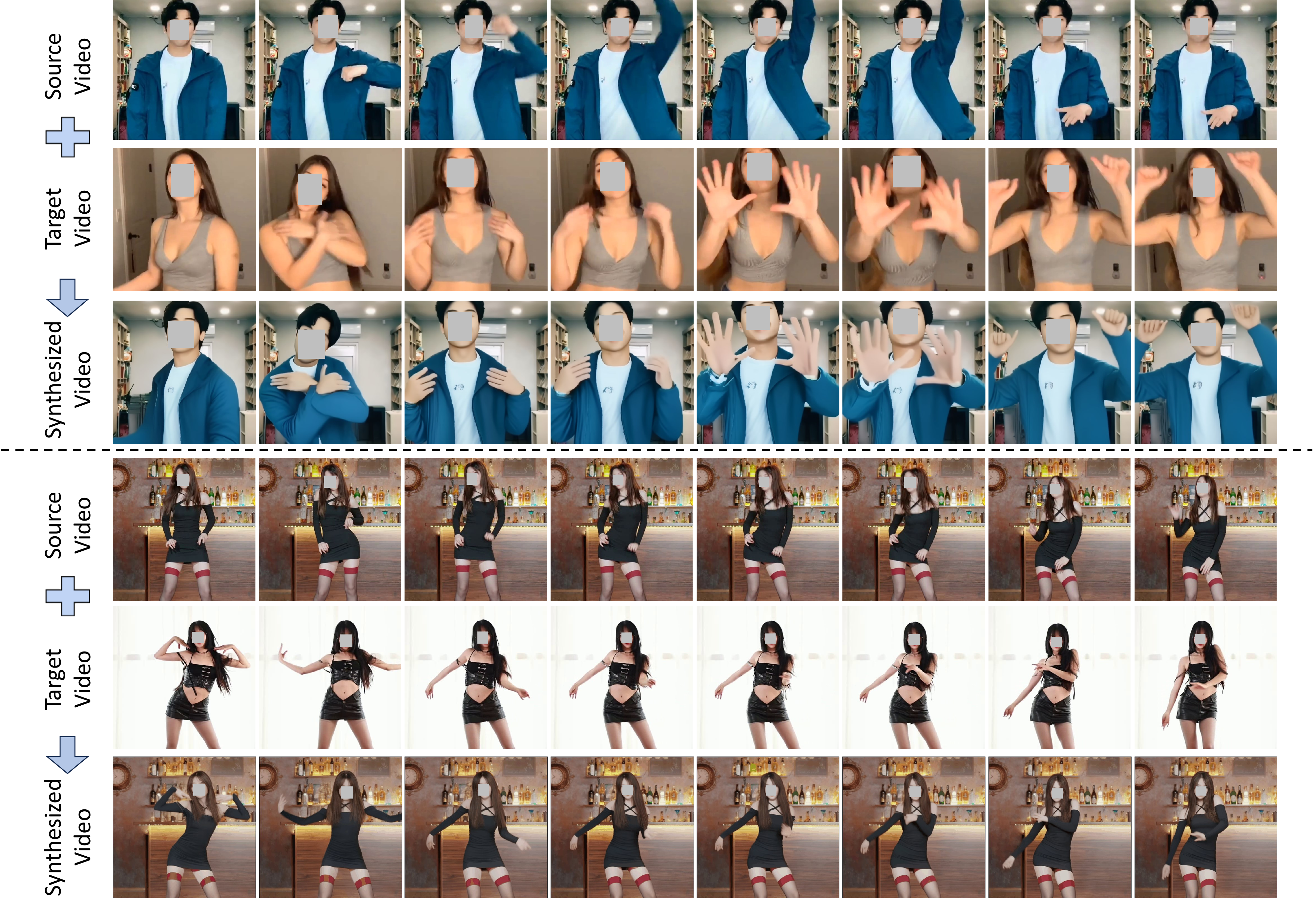}
\end{center}
\vspace{-0.0cm}
   \caption{\small Additional video motion editing results of our MotionFollower (1/4).
   }
\label{fig:supp-bg-1}
\vspace{0.5cm}
\end{figure}

\begin{figure}[H]
\begin{center}
\includegraphics[width=1\linewidth]{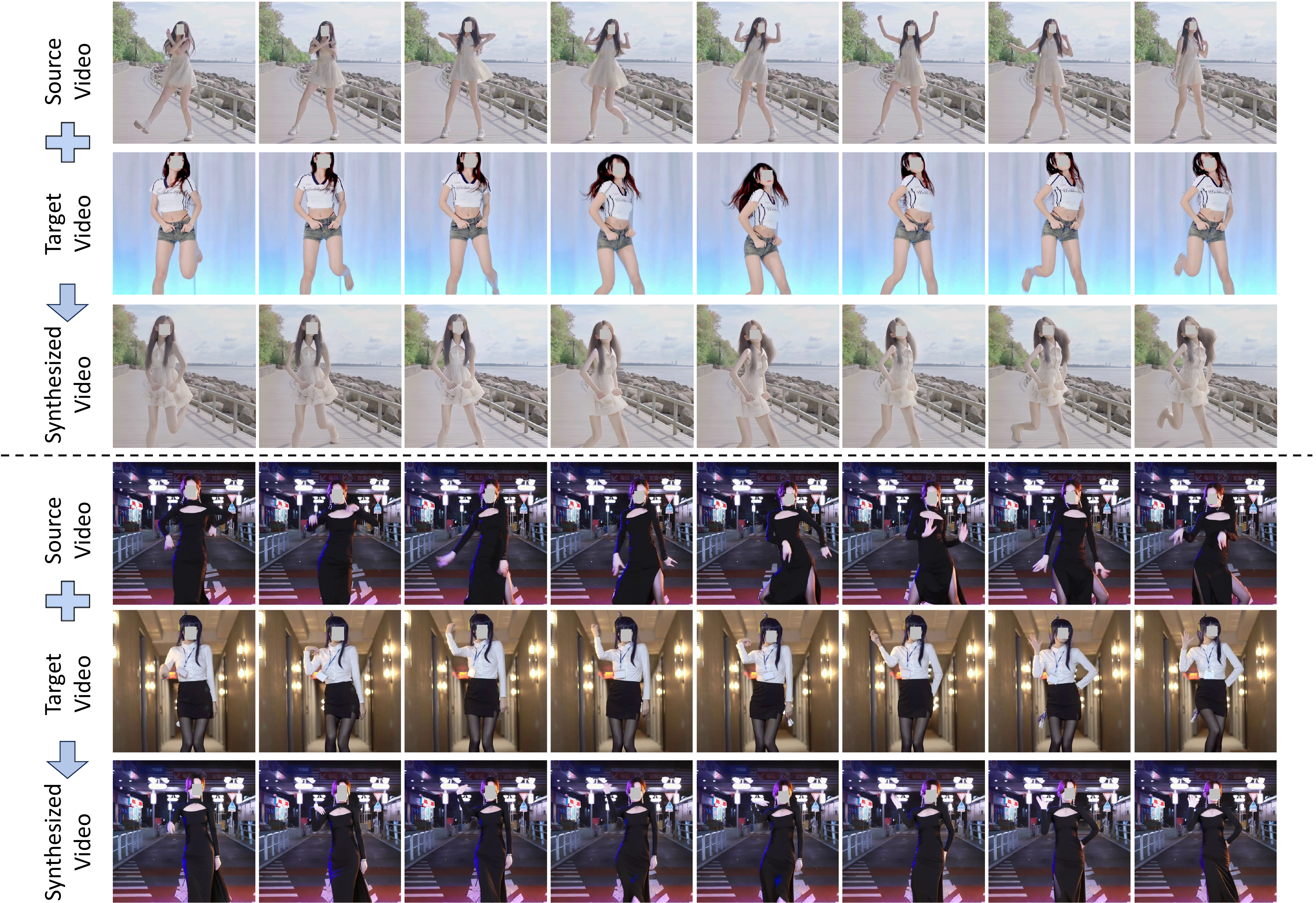}
\end{center}
\vspace{-0.0cm}
   \caption{\small Additional video motion editing results of our MotionFollower (2/4).
   }
\label{fig:supp-bg-2}
\vspace{-0.0cm}
\end{figure}

\begin{figure}[H]
\begin{center}
\includegraphics[width=1\linewidth]{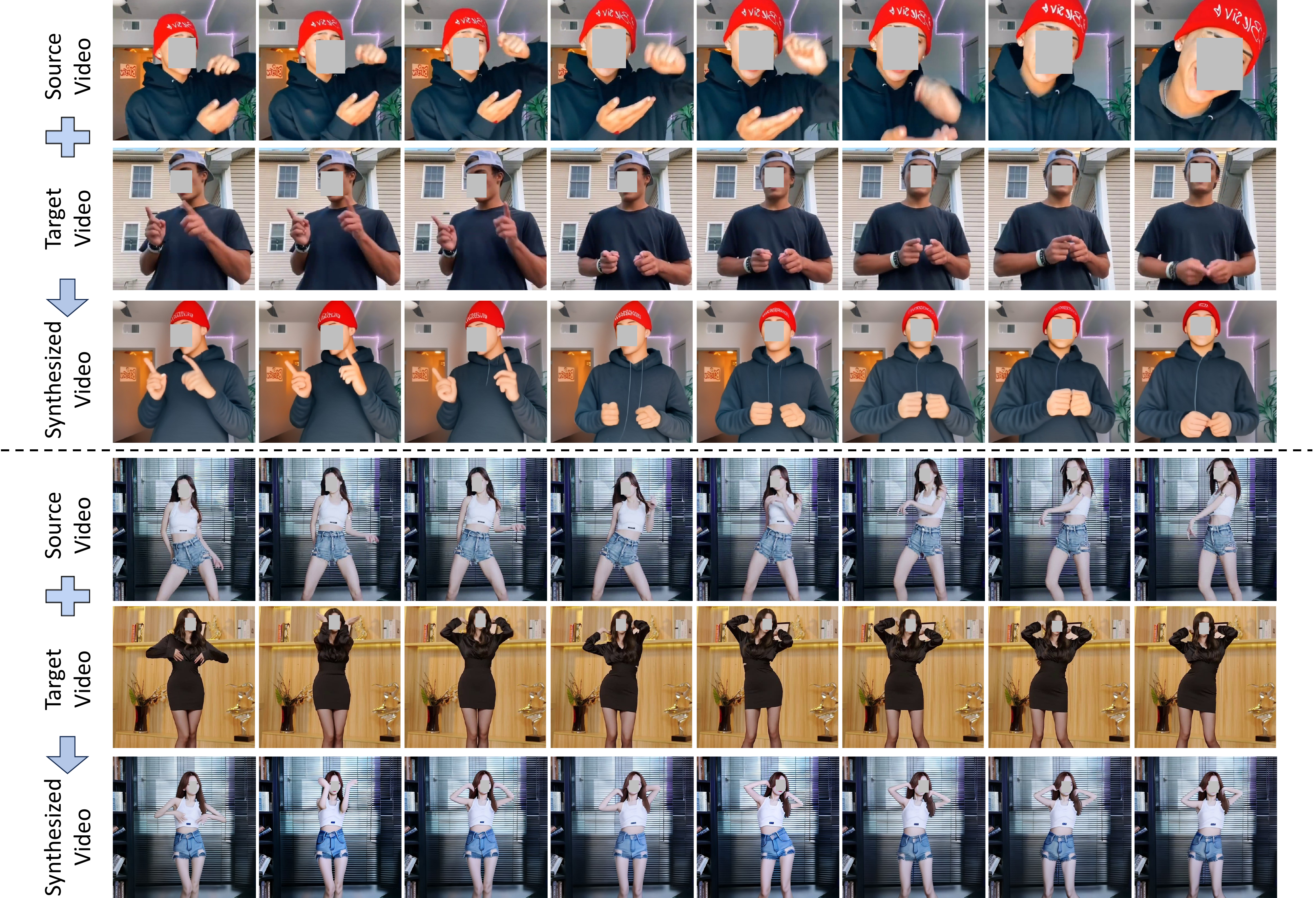}
\end{center}
\vspace{-0.0cm}
   \caption{\small Additional video motion editing results of our MotionFollower (3/4).
   }
\label{fig:supp-pose}
\vspace{0.5cm}
\end{figure}

\begin{figure}[H]
\begin{center}
\includegraphics[width=1\linewidth]{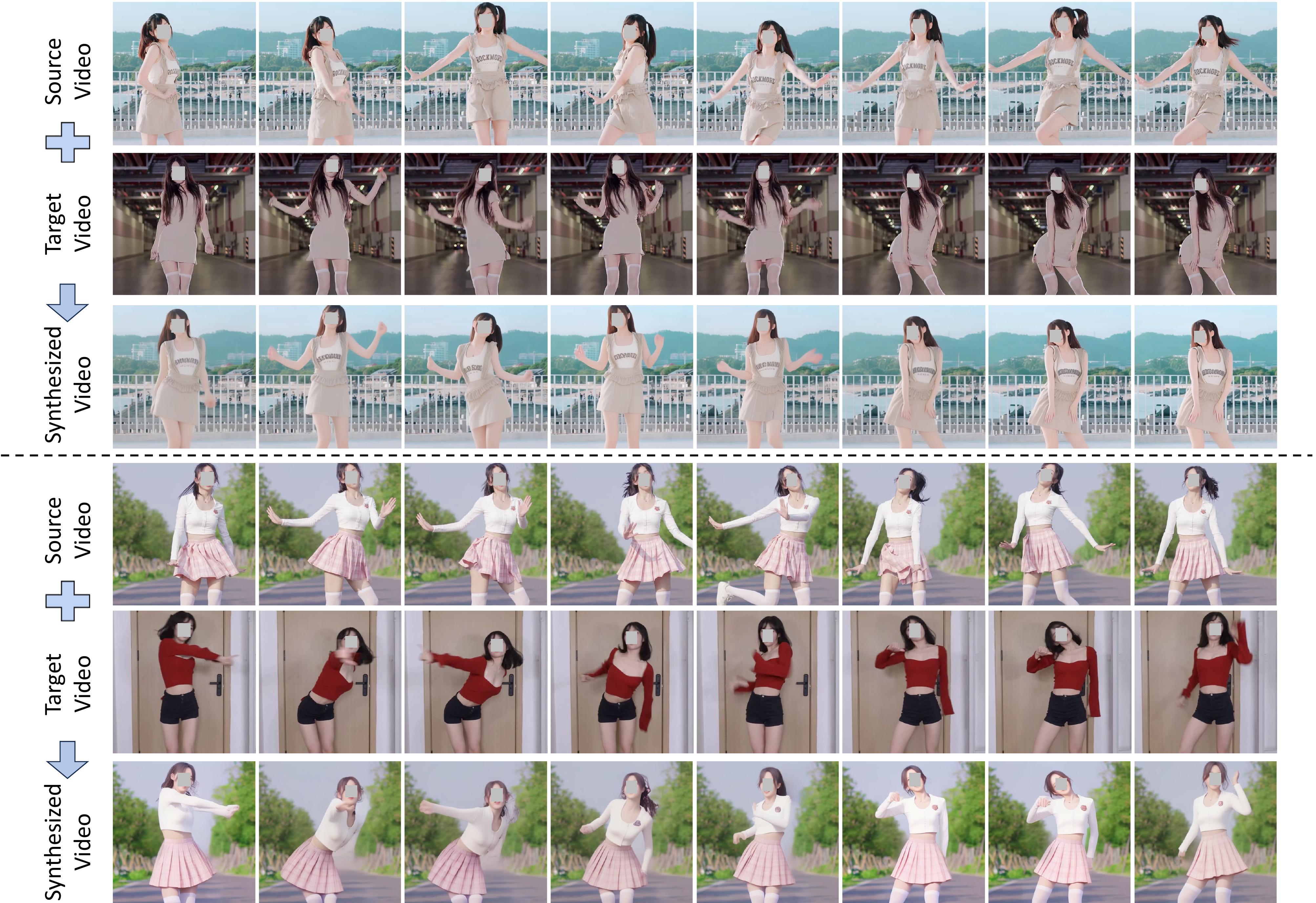}
\end{center}
\vspace{-0.0cm}
   \caption{\small Additional video motion editing results of our MotionFollower (4/4).
   }
\label{fig:supp-camera}
\vspace{-0.0cm}
\end{figure}

Figure \ref{fig:editing_result} illustrates 4 examples of motion editing using MotionFollower.
Figure \ref{fig:supp-bg-1} and Figure \ref{fig:supp-bg-2} show additional video motion editing results of MotionFollower in the specific videos featuring complicated backgrounds. Figure \ref{fig:supp-pose} illustrates video motion editing results of our proposed model in the videos including complex initial poses. Figure \ref{fig:supp-camera} shows the editing results of our MotionFollower in the videos with camera movements.

\begin{figure}[H]
\begin{center}
\includegraphics[width=1\linewidth]{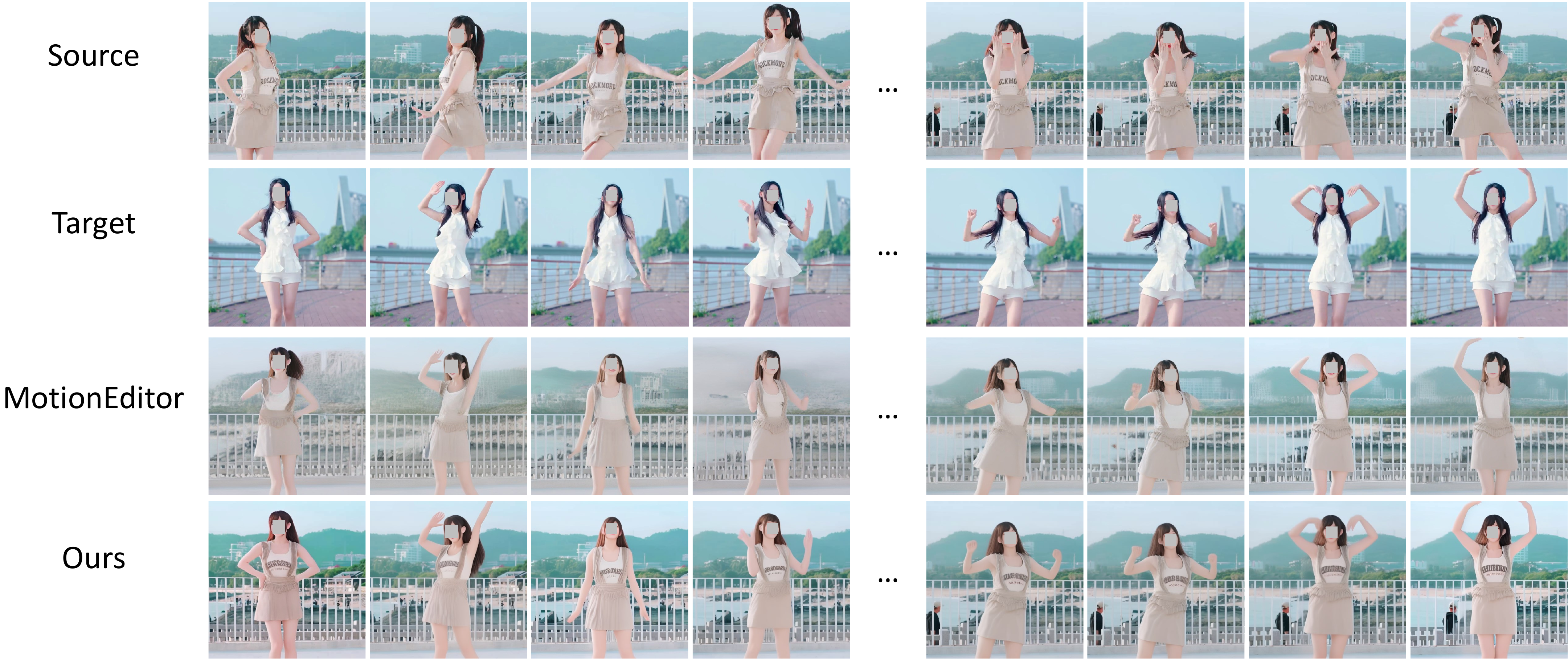}
\end{center}
\vspace{-0.45cm}
   \caption{\small Qualitative comparison on a long video containing complicated appearances and camera movements.
   }
\label{fig:long_video}
\vspace{-0.1cm}
\end{figure}

\begin{figure}[H]
\begin{center}
\includegraphics[width=1\linewidth]{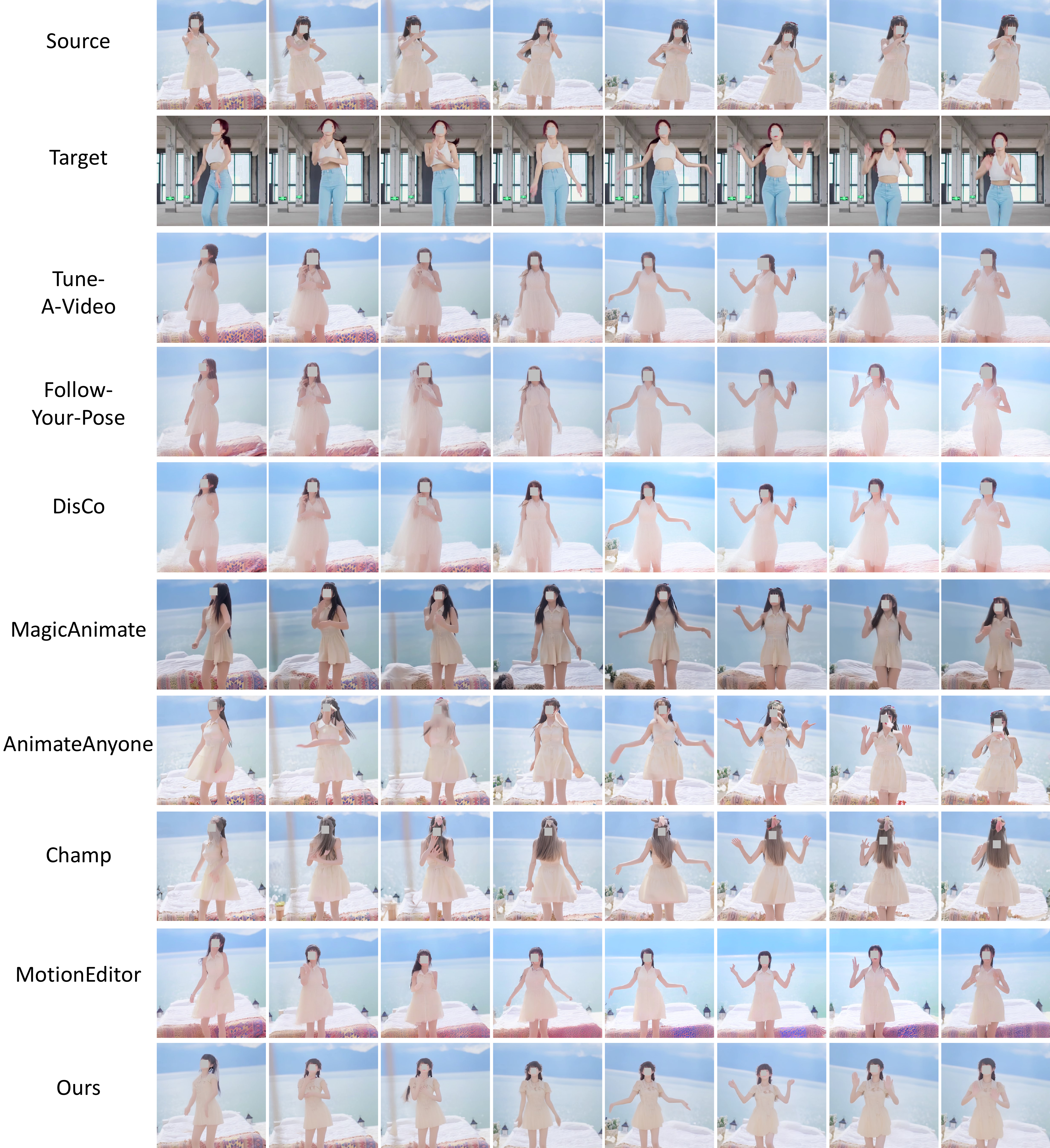}
\end{center}
\vspace{-0.45cm}
   \caption{\small Performance comparison on video containing large camera movements.
   }
\label{fig:camera_movement}
\vspace{-0.2cm}
\end{figure}

\begin{figure}[h!]
\begin{center}
\includegraphics[width=1\linewidth]{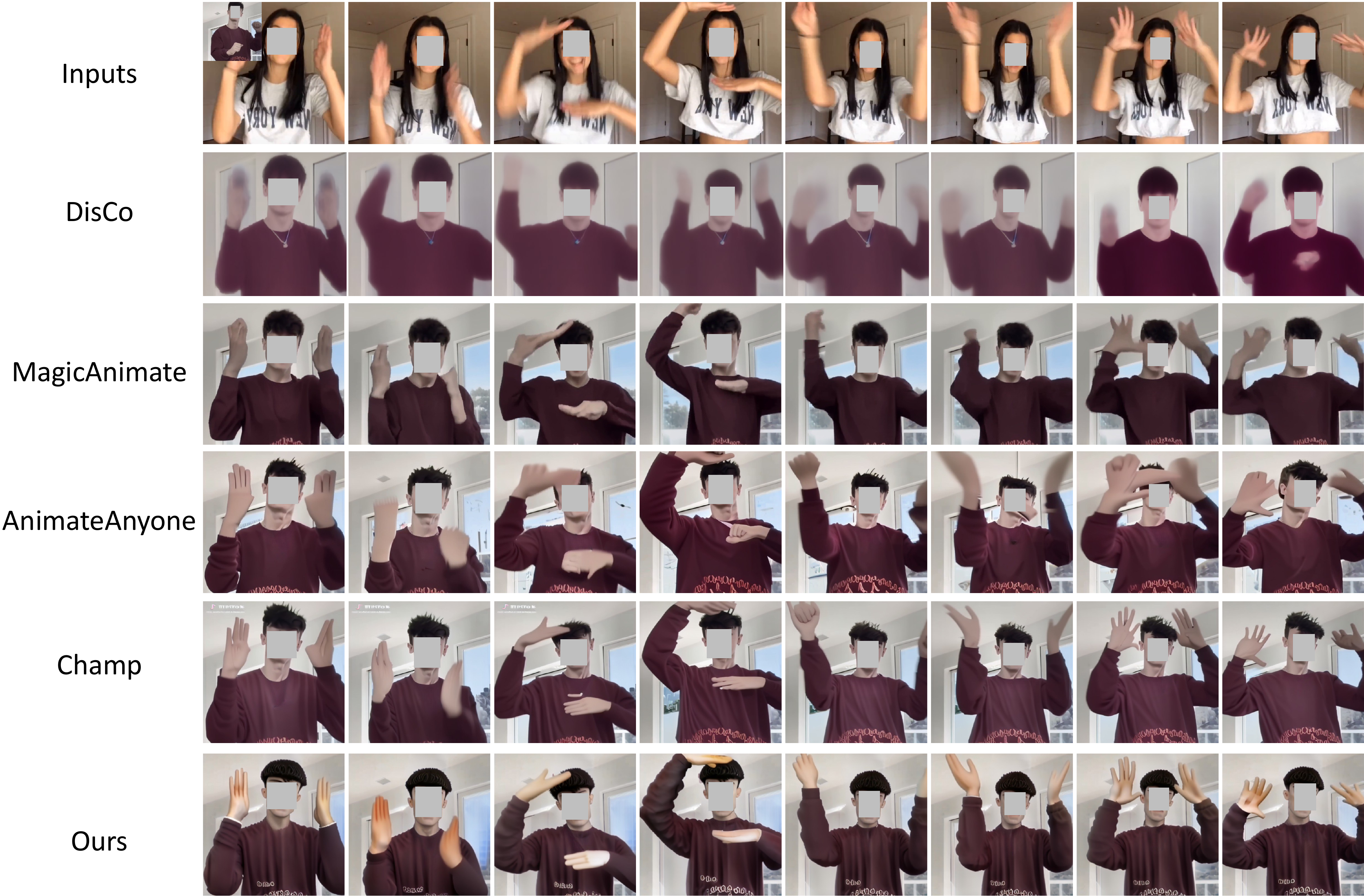}
\end{center}
\vspace{-0.0cm}
   \caption{\small Qualitative comparison between our MotionFollower and other state-of-the-art models in terms of the human motion transfer task.
   }
\label{fig:animation}
\vspace{2.5cm}
\end{figure}

\begin{figure}[h!]
\begin{center}
\includegraphics[width=1\linewidth]{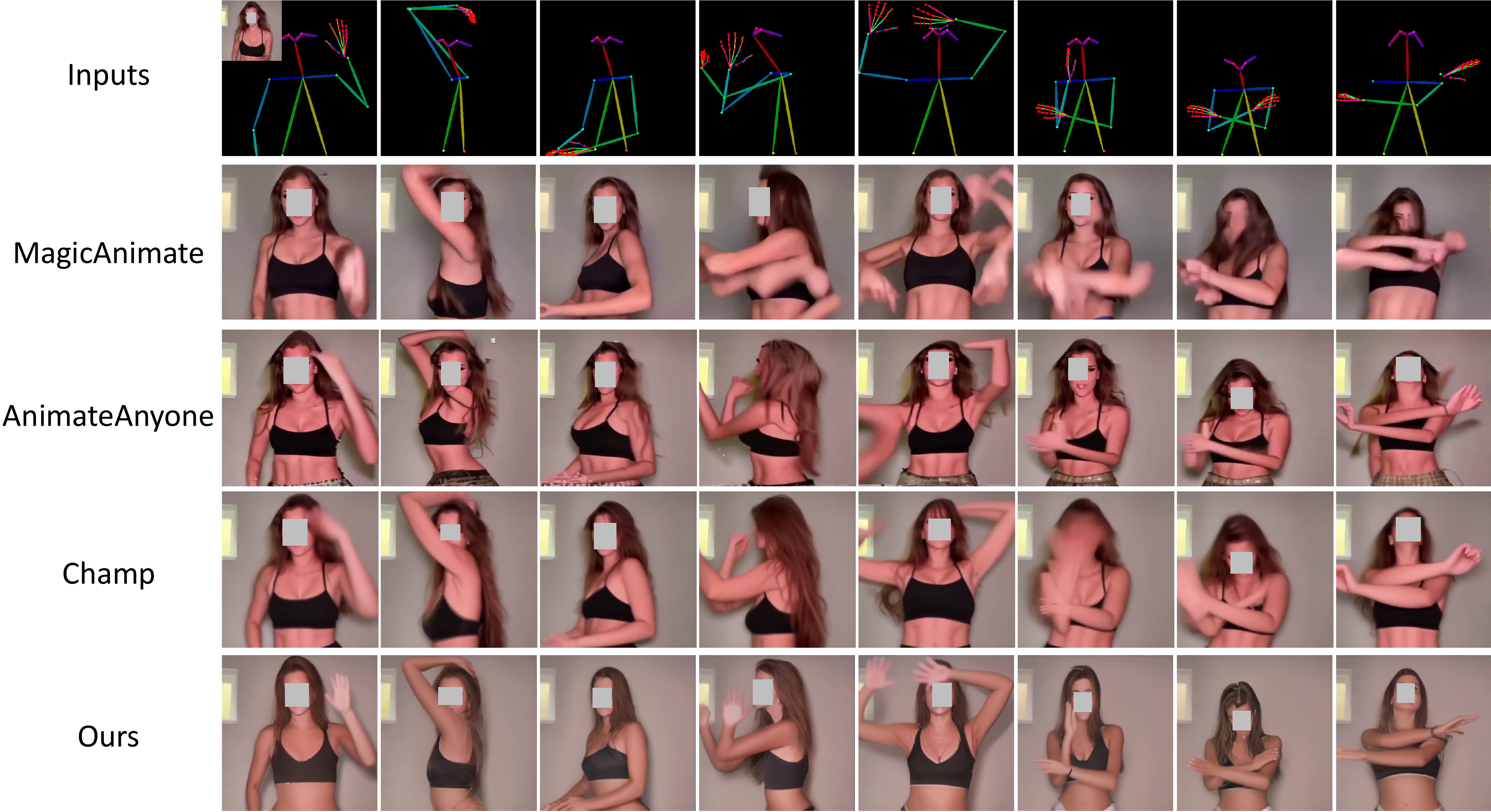}
\end{center}
\vspace{-0.0cm}
   \caption{\small Additional comparison results between our MotionFollower and other models in terms of human motion transfer, using the same case as in \cite{zhu2024champ}.
   }
\label{fig:animation-plus}
\vspace{-0.0cm}
\end{figure}

Figure \ref{fig:long_video} shows the comparison between ours and the strongest competitor MotionEditor on a long video of 600 frames.
Figure \ref{fig:camera_movement} demonstrates the comparisons on a video containing large camera movements.
Figure \ref{fig:animation} presents the comparisons between our method and other approaches to the Human Motion Transfer task.
We also conduct an additional experiment on Human Motion Transfer, using the same case reported in previous human motion transfer papers~\cite{hu2023animate,xu2023magicanimate,zhu2024champ}, as shown in Figure \ref{fig:animation-plus}.

\vspace{-0.5cm}
\begin{table}[t]\small
\caption{\small User preference ratio of MotionFollower when comparing with other competitors. Higher indicates the users prefer more to our MotionFollower.
}
\vspace{-0.0in}
\begin{center}
\renewcommand\arraystretch{1.1}
\scalebox{0.9}{
\begin{tabular}{lccc}
\toprule
Model            & Motion Alignment    & Appearance Alignmen    & Background Alignment    \\ \midrule
MRAA~\cite{siarohin2021motion}             & 95.9\% & 98.8\% & 94.5\% \\
Tune-A-Video~\cite{wu2023tune}     & 94.7\% & 97.2\% & 93.6\% \\
Follow-Your-Pose~\cite{ma2023follow} & 95.5\% & 96.1\% & 96.3\% \\
DisCo~\cite{wang2023disco}            & 92.3\% & 94.4\% & 95.6\% \\
MagicAnimate~\cite{xu2023magicanimate}     & 90.1\% & 91.8\% & 93.3\% \\
AnimateAnyone~\cite{hu2023animate}    & 93.2\% & 93.8\% & 94.7\% \\
Champ~\cite{zhu2024champ}            & 92.0\% & 92.5\% & 91.4\% \\
MotionEditor~\cite{tu2024motioneditor}     & 88.3\% & 85.5\% & 89.7\% \\ \bottomrule
\end{tabular}
}
\end{center}
\label{table:user_study}
\vspace{-0cm}
\end{table}

\vspace{-0cm}

\begin{figure}[H]
\begin{center}
\includegraphics[width=1\linewidth]{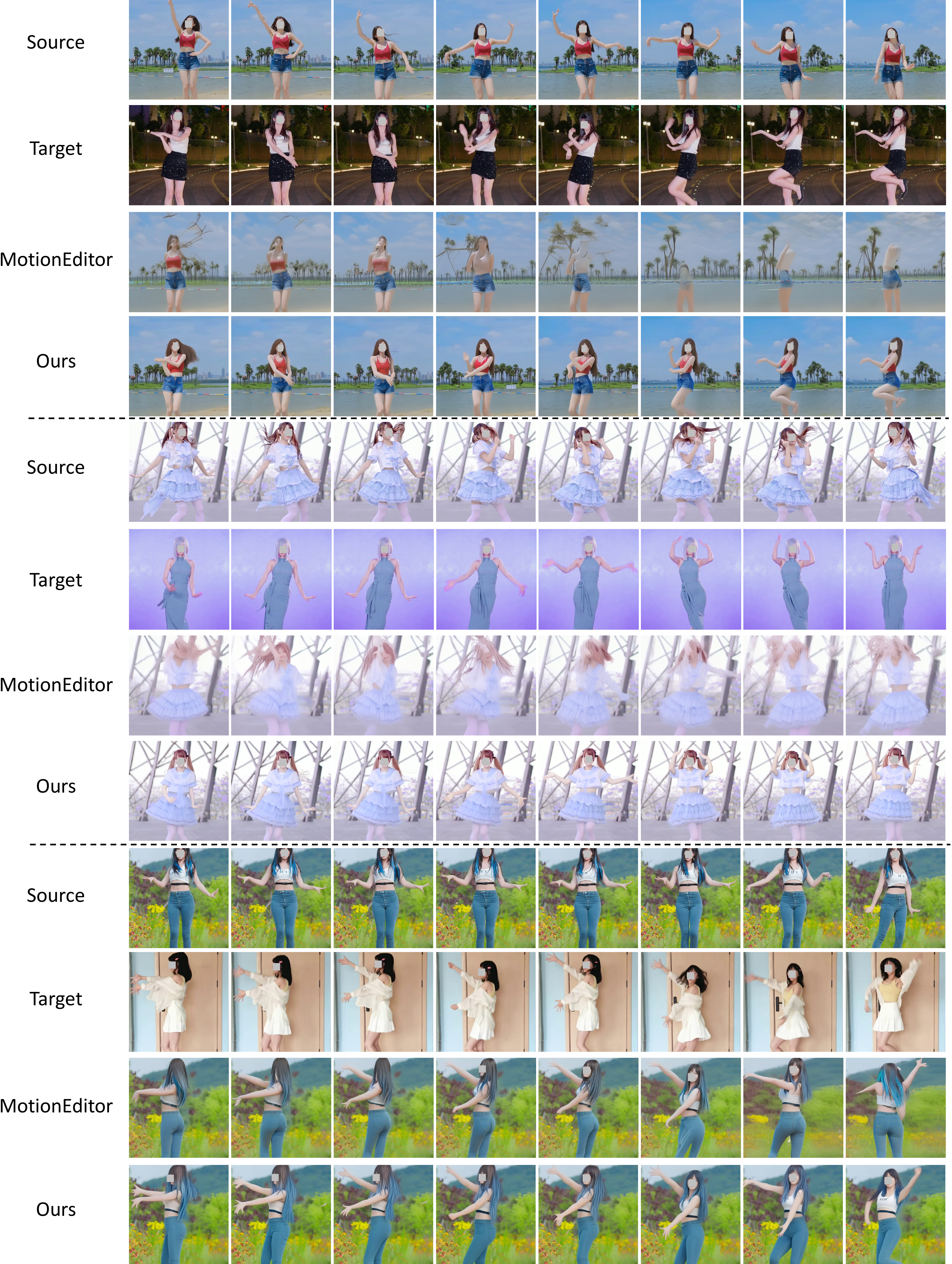}
\end{center}
\vspace{-0.0cm}
   \caption{\small Additional comparison results between our MotionFollower and the strongest competitor MotionEditor.
   }
\label{fig:comparison-plus}
\vspace{1.5cm}
\end{figure}

Additionally, Figure \ref{fig:comparison-plus} shows the additional performance comparison results between our MotionFollower and the most advanced video motion editing model MotionEditor.

\vspace{0.7cm}
\begin{figure}[h!]
\begin{center}
\includegraphics[width=1\linewidth]{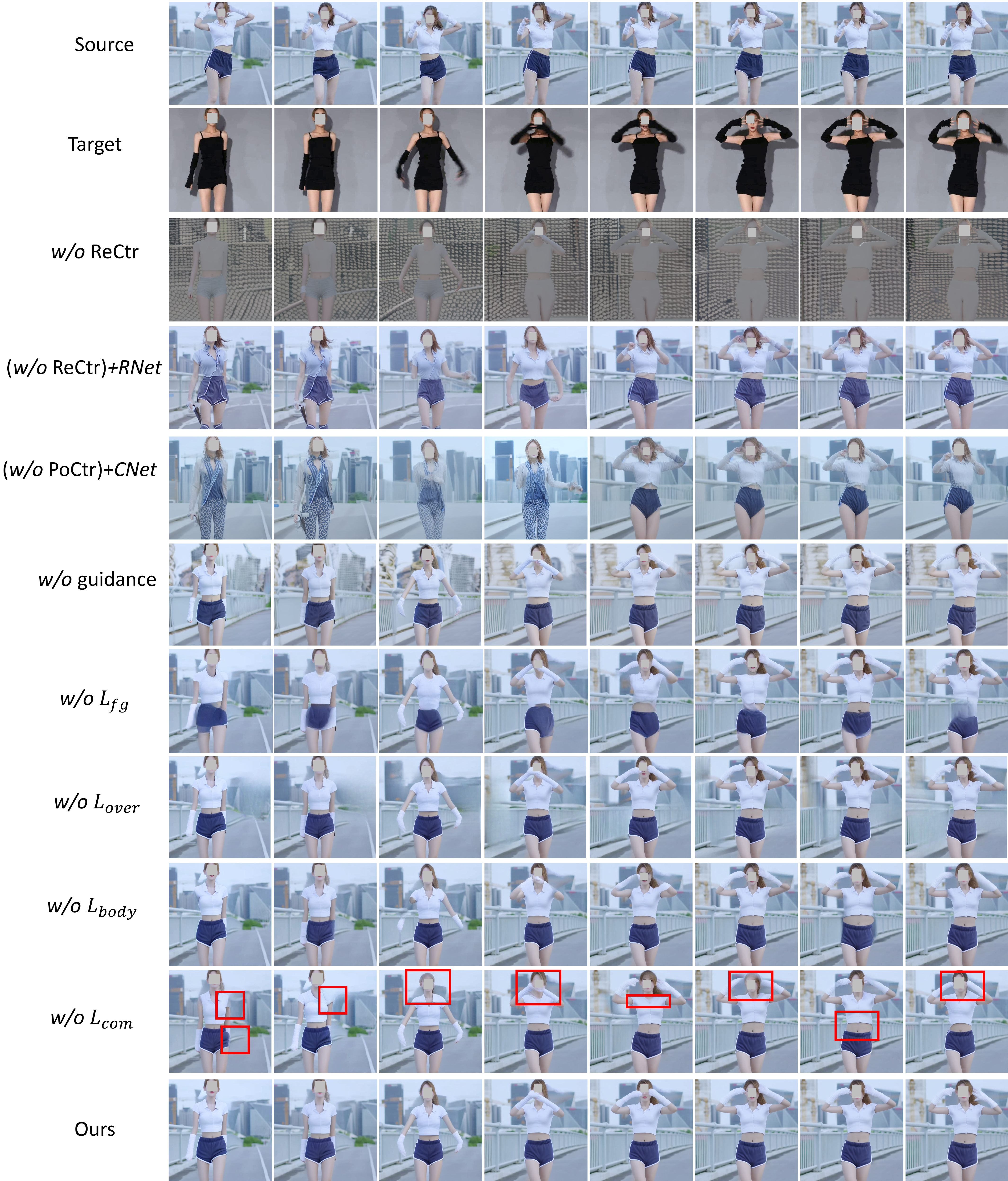}
\end{center}
\vspace{0.0cm}
   \caption{\small A more comprehensive ablation study result.
   }
\label{fig:ablation-plus}
\vspace{1.5cm}
\end{figure}

\subsection{Additional Ablation Study}
\label{sec:additional_ablation}

We conduct a more comprehensive ablation study, as illustrated in Figure \ref{fig:ablation-plus}. We can see that \textit{w/o} $\bm{L}_{fg}$ results in degrading quality of the foreground. The plausible reason is that the diffusion model struggles to preserve the details of the protagonist's appearance without explicit foreground-related guidance. 
\textit{w/o} $\bm{L}_{over}$ contributes to the occurrence of blurry noises in the dynamic background. The main reason is that it is relatively difficult for the diffusion model to capture and model the dynamic background, as its data distribution frequently varies.
\textit{w/o} $\bm{L}_{body}$ and \textit{w/o} $\bm{L}_{com}$ both cause some semantic distortion due to the interference from non-overlapping protagonist's parts.
The results demonstrate that our proposed loss function can effectively promote the diffusion model to preserve the appearance of foregrounds and backgrounds.

\subsection{Limitations}
\label{sec:limitations}

\begin{figure}[h!]
\begin{center}
\includegraphics[width=0.9\linewidth]{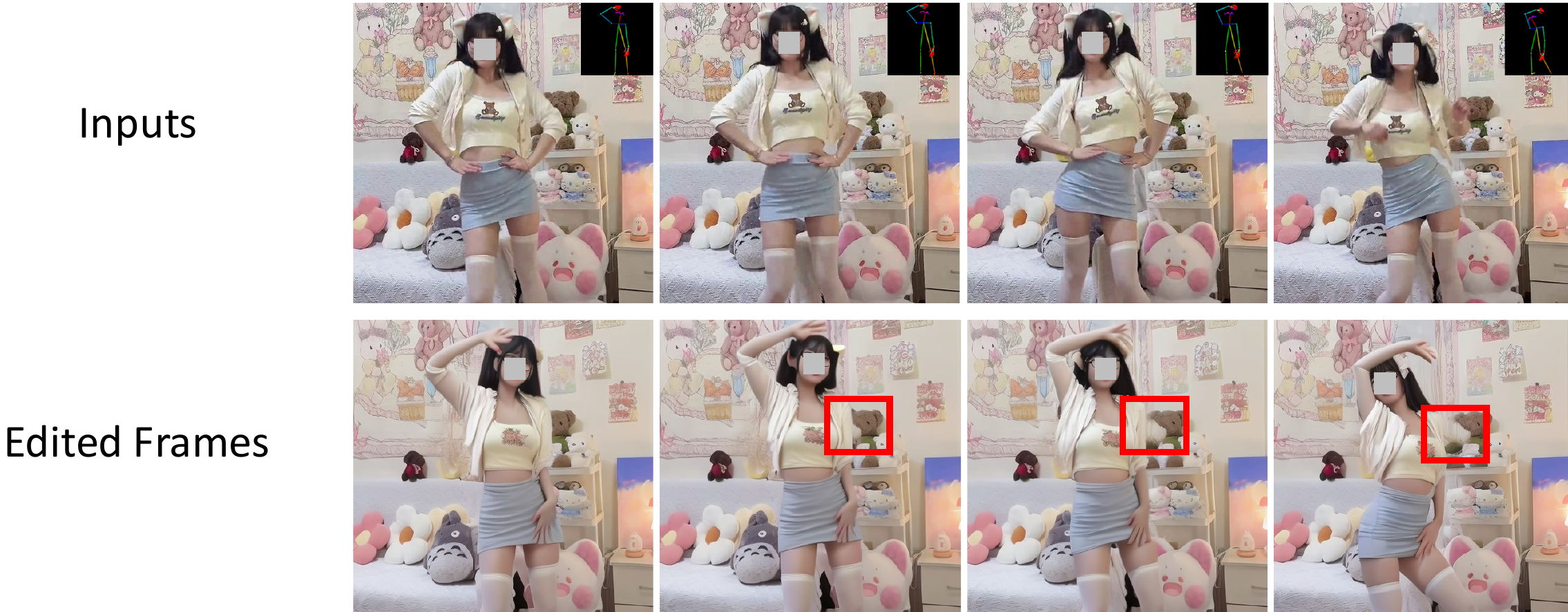}
\end{center}
\vspace{-0.2cm}
   \caption{\small One failure case of our MotionFollower.
   }
\label{fig:limitation}
\vspace{-0.2cm}
\end{figure}

Figure \ref{fig:limitation} shows one failure case of our MotionFollower. The toy bear located behind the girl remains incomplete due to the foreground obstruction in the source video. Our model struggles to fill in the obscured areas in the background when encountering numerous distinct small objects in the background. The probable solution is to explicitly introduce an additional inpainting adaptor to the diffusion model for recovering the background areas. This part is left as future work.

\subsection{Ethical Concern}
\label{sec:ethical_concern}

While MotionFollower has broad applicability, it is crucial to address several concerns: the risks of misuse in creating deceptive media, potential biases in training data, and the importance of respecting intellectual property.

\FloatBarrier 

\end{document}